\documentclass[acmtog]{acmart}

\usepackage{booktabs} 
\usepackage{graphicx}
\usepackage{enumitem}
\usepackage{multirow} 
\usepackage{subfig} 
\usepackage{cuted}

\usepackage{tcolorbox}

\definecolor{colorcommentfg}{RGB}{0,63,87}
\definecolor{colorcommentbg}{RGB}{255,255,255}
\definecolor{colorcommentframe}{RGB}{0,0,0}
\usepackage[ruled]{algorithm2e} 

\SetAlFnt{\small}
\SetAlCapFnt{\small}
\SetAlCapNameFnt{\small}
\SetAlCapHSkip{0pt}

\definecolor{gre}{HTML}{39b54a}
\begin{document}
\thispagestyle{empty}
\title{Divide and Conquer: Language Models can Plan and Self-Correct for Compositional Text-to-Image Generation}


\author{Zhenyu Wang}
\affiliation{%
 \institution{Tsinghua University}
 \city{Beijing}
 \country{China}}
\email{wangzy20@mails.tsinghua.edu.cn}
\author{Enze Xie$^*$}
\affiliation{%
 \institution{Noah's Ark Lab, Huawei}
 \city{Hong Kong}
 \country{China}
}
\email{xieenze@connect.hku.hk}
\author{Aoxue Li}
\affiliation{%
 \institution{Noah's Ark Lab, Huawei}
 \city{Beijing}
 \country{China}
}
\email{lax@pku.edu.cn}
\author{Zhongdao Wang}
\affiliation{%
 \institution{Noah's Ark Lab, Huawei}
 \city{Beijing}
 \country{China}
}
\email{wangzhongdao@huawei.com}
\author{Xihui Liu}
\affiliation{%
 \institution{The University of Hong Kong}
 \city{Hong Kong}
 \country{China}
}
\email{xihuiliu@eee.hku.hk}
\author{Zhenguo Li}
\affiliation{%
 \institution{Noah's Ark Lab, Huawei}
 \city{Hong Kong}
 \country{China}
}
\email{Li.Zhenguo@huawei.com}

\thanks{Project page: {\color{gray} \url{https://zhenyuw16.github.io/CompAgent/}} }

\authorsaddresses{$^*$: corresponding author. \\ 
Mails: wangzy20@mails.tsinghua.edu.cn; xieenze@connect.hku.hk; lax@pku.edu.cn; wangzhongdao@huawei.com; xihuiliu@eee.hku.hk; Li.Zhenguo@huawei.com}

\settopmatter{printacmref=false}
\setcopyright{none}
\renewcommand\footnotetextcopyrightpermission[1]{}
\pagestyle{empty}
 \fancyfoot[RO,LE]{}
\settopmatter{printfolios=true}

\begin{abstract}

Despite significant advancements in text-to-image models for generating high-quality images, these methods still struggle to ensure the controllability of text prompts over images in the context of complex text prompts, especially when it comes to retaining object attributes and relationships. In this paper, we propose \textit{CompAgent}, a \textit{training-free} approach for compositional text-to-image generation, with a large language model (LLM) agent as its core. The fundamental idea underlying CompAgent is premised on a divide-and-conquer methodology. Given a complex text prompt containing multiple concepts including objects, attributes, and relationships, the LLM agent initially decomposes it, which entails the extraction of individual objects, their associated attributes, and the prediction of a coherent scene layout. These individual objects can then be independently conquered. Subsequently, the agent performs reasoning by analyzing the text, plans and employs the tools to compose these isolated objects. The verification and human feedback mechanism is finally incorporated into our agent to further correct the potential attribute errors and refine the generated images. Guided by the LLM agent, we propose a tuning-free multi-concept customization model and a layout-to-image generation model as the tools for concept composition, and a local image editing method as the tool to interact with the agent for verification. The scene layout controls the image generation process among these tools to prevent confusion among multiple objects. 
Extensive experiments demonstrate the superiority of our approach for compositional text-to-image generation: CompAgent achieves more than 10\% improvement on T2I-CompBench, a comprehensive benchmark for open-world compositional T2I generation. The extension to various related tasks also illustrates the flexibility of our CompAgent for potential applications.
\end{abstract}

\ccsdesc[500]{Computing methodologies~Computer vision}
\ccsdesc[500]{Computing methodologies~Image manipulation}

\keywords{Image Generation, Compositional Text-to-Image Generation, Diffusion Models, LLM Agent}

\begin{teaserfigure}
\setlength{\abovecaptionskip}{0pt}
\setlength{\belowcaptionskip}{0pt}
\centering
  \includegraphics[width=0.91\textwidth]{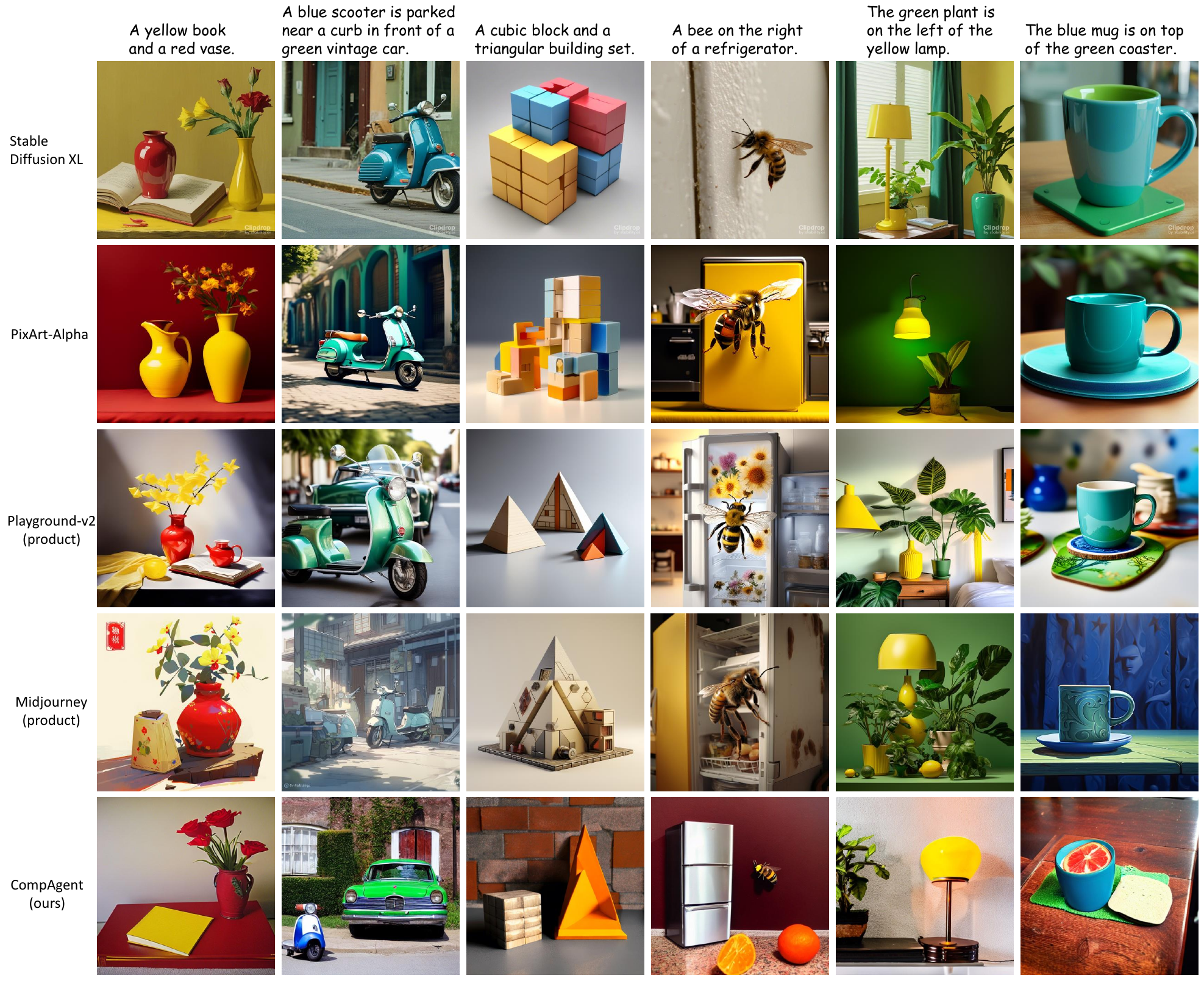}
  \caption{For compositional text-to-image generation,  existing state-of-the-art text-to-image models often fail to generate images that accurately correspond to the description of text inputs, especially about object attributes and relationships. In comparison, our method manages to generate images with correct objects and attributes according to the complex text prompts. }
  \label{fig:comparison1}
\end{teaserfigure}

\maketitle

\section{Introduction}

Recent advancements in text-to-image generation~\cite{rombach2022high, saharia2022photorealistic, ramesh2022hierarchical, chang2023muse, chen2023pixart} have demonstrated the remarkable capability in creating diverse and high-quality images based on language text inputs. However, even state-of-the-art text-to-image models often fail to generate images that accurately align with complex text prompts, where multiple objects with different attributes or relationships are composed into one scene~\cite{feng2023trainingfree, huang2023t2i, chefer2023attend}. As can be seen in Fig.~\ref{fig:comparison1}, existing methods cannot create correct objects, attributes, or relationships within the generated images given these complex text prompts.

Addressing compositional text-to-image generation requires solving at least the following three issues: 1) \textit{Object types and quantities.} Due to the presence of multiple objects, the generated images should accurately incorporate each object, avoiding issues such as incorrect object types, omissions of objects, and discrepancies in object quantities. 2) \textit{Attribute binding.} Objects inherently possess distinctive attributes, like "color", "shape" or "texture". It should be guaranteed that these object attributes are meticulously preserved within the generated images, avoiding issues like attribute misalignment or leakage. 3) \textit{Object relationships.} There can be interaction relationships among the multiple objects, such as spatial relationships like "left", "right" or non-spatial ones like "holding", "playing". The generation process should encapsulate and convey these relationships within the resultant images with precision and fidelity.

\begin{figure}
      \subfloat[]{ \includegraphics[width=0.98\columnwidth]{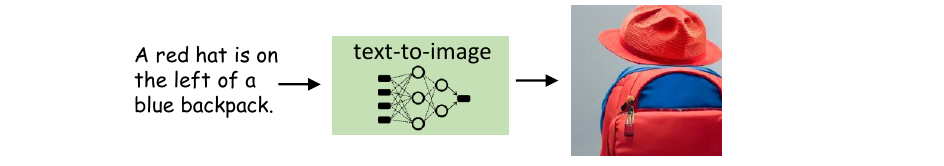} \label{fig:frame1}}\\
      \subfloat[]{ \includegraphics[width=0.98\columnwidth]{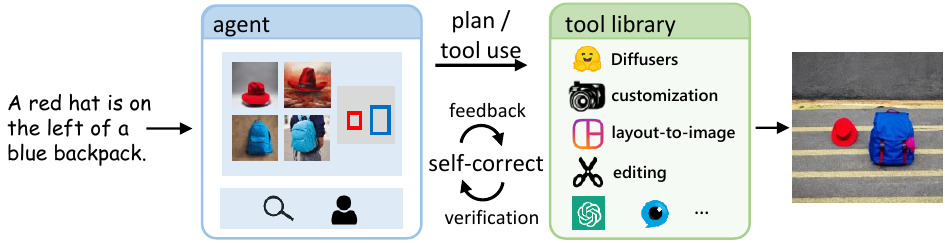} \label{fig:frame2}} \\
\caption{\textbf{The overview of existing text-to-image generation methods (a) and our CompAgent (b).} Existing methods generate images from text prompts in a single step. In comparison, guided by the LLM agent, CompAgent divides complex text prompts into individual objects, conquers them separately, and composes them into final images with the tool library.}
\label{fig:frame}
\end{figure}

While existing text-to-image models do not possess the capability to address the aforementioned three issues, they do demonstrate proficiency in the generation of single objects, encompassing their distinctive types and attributes. Motivated by this, we propose \textit{CompAgent}, a training-free approach founded upon the principle of divide-and-conquer for compositional text-to-image generation, coordinated by a large language model (LLM) agent. The fundamental idea revolves around breaking down intricate textual sentences into their constituent individual objects, initially ensuring the correctness of these isolated objects, then composing them together to produce the final images. The overview is illustrated in Fig.~\ref{fig:frame}. %

The core of our method is an AI agent implemented by LLM, which serves as the "brain" of the framework and is primarily responsible for the following tasks: 1) \textit{Decomposition.} The agent decomposes the complex compositional sentence,  extracting and cataloging all objects and their associated attributes. Simultaneously, it designs the layout of the scene, defining the positions of the objects through the specification of bounding boxes. The text-to-image generation models are then engaged to generate images for each individual object. 2) \textit{Planning and Tool use.} The LLM agent conducts reasoning according to the complex text prompt and then formulates a strategic approach to image generation that is contingent upon the presence of object attributes and relationships. Then it employs external tools to perform image generation or editing. 3) \textit{Verification and Feedback.} By leveraging the vision ability of LLM or other visual models, the agent further scrutinizes the generated images, discerns and rectifies potential attribute errors. Additionally, human feedback can be seamlessly incorporated into scene layout refinement, thereby enhancing the quality of the ultimate outputs.

Guided by the LLM agent, we introduce three tools to compose multiple individual objects into a single cohesive image according to the scene layout. The first is about \textit{tuning-free multi-concept customization}. Specifically, we impose spatial layout constraints by cross-attention map editing and a pre-trained ControlNet~\cite{zhang2023adding} on a tuning-free single-concept customization network~\cite{li2023blip} for supporting multiple objects. It regards previously generated images of individual objects as user-specified subjects to create customized images. In this way, the object attributes can be guaranteed. The second is a \textit{layout-to-image} model. Through latent updating~\cite{xie2023boxdiff}, images are generated with the bounding box layout as the condition. The specification of object types and quantities is made feasible through the imposition of layout conditions, enabling the model to place emphasis on the accurate representation of object relationships. However, the layout-to-image model may not necessarily produce object attributes with absolute accuracy. Therefore, we further present the third tool, \textit{local image editing}. The agent examines the objects with attribute errors in the generated images, which will be masked out through a segmentation model~\cite{kirillov2023segment}. Through cross-attention control, subject-driven editing is conducted through the previous multi-concept customization network to replace the erroneous objects with their correct attribute counterparts. In addition to the aforementioned three tools, our toolkit also involves existing text-to-image generation models and vision-language multi-modal models, to handle simple text prompts and assess attribute correctness.

Our main contributions can be summarized as follows:

\begin{itemize}[topsep=0pt, parsep=0pt, itemsep=0pt, partopsep=0pt]
    \item We propose CompAgent to address compositional text-to-image generation through the divide-and-conquer approach. The LLM agent oversees the entire task, performing decomposition, reasoning and overall planning, and tool library use to solve complex cases in text-to-image generation.
    \item By employing both global and local layout control about the spatial arrangement, we propose a tuning-free multi-concept customization model to address the attribute binding problem. We also observe that the layout-to-image generation manner can ensure the faithfulness of object relationships.
    \item We introduce the verification and feedback mechanism into our LLM agent. By interacting with our designed local image editing tool, the potential attribute errors can be corrected and the generated images can be further refined.
\end{itemize}

We evaluate CompAgent on the recent T2I-CompBench benchmark~\cite{huang2023t2i} for compositional text-to-image generation, involving extensive object attributes and relationships. Both qualitative and quantitative results demonstrate the superiority of our method. We achieve more than 10\% improvement in the metrics about compositional text-to-image evaluation. Furthermore, CompAgent can be flexibly extended to various related applications, like multi-concept customized image generation, reference-based image editing, object placement, and so on.

\section{Related Work}

\quad \textit{Text-to-image generation.} With the development of diffusion models~\cite{dhariwal2021diffusion, ho2020denoising}, text-to-image generation has achieved remarkable success~\cite{rombach2022high, saharia2022photorealistic, chen2023pixart}. These models can typically generate highly natural and realistic images, but cannot guarantee the controllability of texts over images. Many subsequent works are thus proposed for controllable text-to-image generation. ControlNet~\cite{zhang2023adding} controls Stable Diffusion with various conditioning inputs like Canny edges, and~\cite{voynov2023sketch} adopts sketch images for conditions. Layout-to-image methods~\cite{li2023gligen, xie2023boxdiff, lian2023llm, chen2024training} synthesize images according to the given bounding boxes of objects. And some image editing methods~\cite{yang2023paint, chen2023anydoor, brooks2023instructpix2pix, parmar2023zero} edit images according to the user's instructions. Despite the success of these methods, they are still limited in handling complex text prompts for image generation.

\textit{LLM agent.} Large language models (LLMs), like ChatGPT, GPT-4~\cite{gpt42023}, Llama~\cite{touvron2023llama, touvron2023llama2}, have demonstrated impressive capability in natural language processing. The involvement of vision ability in GPT-4V~\cite{yang2023dawn} further enables the model to process visual data. Recently, LLMs begin to be adopted as agents for executing complex tasks. These works~\cite{yang2023auto, shen2023hugginggpt, liu2023internchat} apply LLMs to learn to use tools for tasks like visual interaction, speech processing, and more recent works have extended to more impressive applications like software development~\cite{qian2023communicative}, gaming~\cite{meta2022human}, or APP use~\cite{yang2023appagent}. Different from them, we focus on the task of compositional text-to-image generation, and utilize a LLM agent for complex text prompt analysis and method planning.

\begin{figure*}
   \centering
\includegraphics[width=0.9\textwidth]{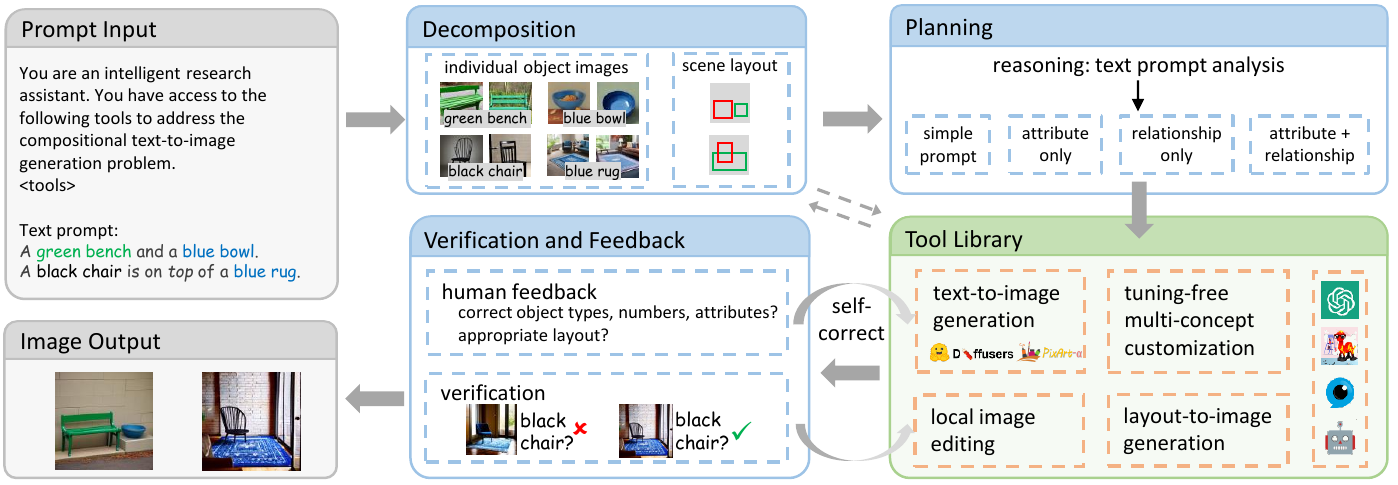}
   \caption{\textbf{The framework of CompAgent.} Given the input containing the complex text prompt, the LLM agent conducts the decomposition and planning tasks to invoke external tools for image generation. It then performs verification or involves human feedback and interacts with the tools for image self-correction. The final image output will well satisfy the requirements from the input text prompt.}
   \label{fig:overview}
\end{figure*}

\section{Method}

The overview of our CompAgent is illustrated in Fig.~\ref{fig:overview}. The LLM agent coordinates the entire framework. It decomposes the complex text prompt, analyzes the attributes within the text, and designs the plans for the tools to be used. Ultimately, it invokes tools to address the challenges inherent in compositional text-to-image generation. 

\subsection{LLM Agent}

The primary responsibilities of our LLM agent center around the execution of tasks including decomposition, planning, tool use, as well as the implementation of verification and feedback.

\textit{Complex text prompt decomposition.} The input text prompts usually contain multiple objects, each characterized by its distinct attributes. These objects may also interact with each other with specific relationships. Existing image synthesis models cannot capture all these attributes and relationships simultaneously, making it difficult to accurately generate images that align with the input texts.

To address the challenges, the LLM agent decomposes the complex texts. It extracts the discrete objects embedded within the text prompts, along with their associated attributes. The text-to-image generation model is then harnessed to generate several images for these objects. Since only one object is involved, existing models can typically generate images that match the attributes. The agent also formulates the scene layout represented as captioned bounding boxes, where the position of each foreground object is specified in the (x, y, width, height) format. The layout will guide the image generation process when composing separate objects together.

\textit{Tool use.} Since the language model does not have the ability for image generation, external tools are required to be utilized to compose separate objects together for image generation:

1) Tuning-free multi-concept customization. It regards previously generated images of individual objects as the target images, and produces images that feature these specific objects according to the input text prompts. For these objects, the presence of multiple images, each characterizing them with accurate attributes, generally assures that the multi-concept customization model can effectively ensure the fidelity of object attributes. However, the model tends to focus excessively on preserving the features of the target images, thus could potentially overlook other information in the texts like object relationships. Therefore, this tool can effectively address the attribute binding problem, but may not guarantee the relationships.

2) Layout-to-image generation. It generates images with the previously established scene layout as the condition, and does not utilize the target images of objects. The object types and quantities can be specified through the scene layout, thereby facilitating the model's enhanced attention to information beyond the objects, \textit{i.e.} the object relationships. As a result, such a layout-to-image generation tool can address the object relationship problem well, but the layout guidance only is not enough to guarantee the object attributes.

3) Local image editing. Since the layout-to-image generation model may not consistently  produce objects with correct attributes, we further design the local image editing tool to replace the object with the correct attribute one. Previously generated images of individual objects are leveraged here as the reference for object replacement. This tool will interact reciprocally with the verification mechanism of the agent, collaboratively determining which object requires to be modified.

4) Text-to-image generation. Existing text-to-image generation models are utilized to generate individual object images during the decomposition step. They will also be used as the tool to generate images given simple text prompts for non-compositional generation.

5) Vision-text multi-modal model. It assesses whether the object attributes in the images are correct for the verification function by leveraging the visual question answering ability of these models.

\textit{Planning.} In light of the diverse proficiency exhibited by the tools, the strategic selection of tool deployment is a core responsibility of our LLM agent. It mainly analyzes the input text prompt. If the text primarily centers around object attributes and the relationships are relatively straightforward (like spatial relationships "left", "right"), the customization tool will be employed. If object relationships are contained while their attributes are simple (like the naive "white" snow color attribute), the layout-to-image generation tool is suitable. If the text involves both attributes and relationships, the LLM agent first employs the layout-to-image model to represent object relationships. Then it will leverage the vision-text multi-modal models to scrutinize the correctness of object attributes for verification, and decide whether to adopt the local image editing tool. For simple text prompts with only straightforward attributes or relationships, the text-to-image generation tool will be directly utilized.

\textit{Verification and feedback.} Considering the potential limitation of the layout-to-image generation tool in accurately rendering object attributes, a verification process becomes imperative. We employ existing vision-language multi-modal models like GPT-4V~\cite{yang2023dawn}, MiniGPT-4~\cite{zhu2023minigpt}, LLaVA~\cite{liu2023visual}, and query it about whether the attributes are correct. If the attributes of some objects are incorrect, the LLM agent will invoke the local image editing tool to substitute the objects with the correct ones. 

Besides, for too complicated text prompts with a higher number of objects, together with intricate attributes and relationships, relying solely on the agent to automatically decompose the text and design scene layout may not necessarily be entirely accurate. In this situation, human feedback can be involved. Humans can make adjustments to the scene layout, such as inappropriate object sizes, positions, missing or extra objects. They can also make modifications to planning and verification errors. Introducing human feedback in the form of layout can reduce the cost of human labor. The accommodation of our framework about human feedback makes our LLM agent more flexible for compositional generation.

\subsection{Tuning-Free Multi-Concept Customization}

\begin{figure}
   \centering
\includegraphics[width=\columnwidth]{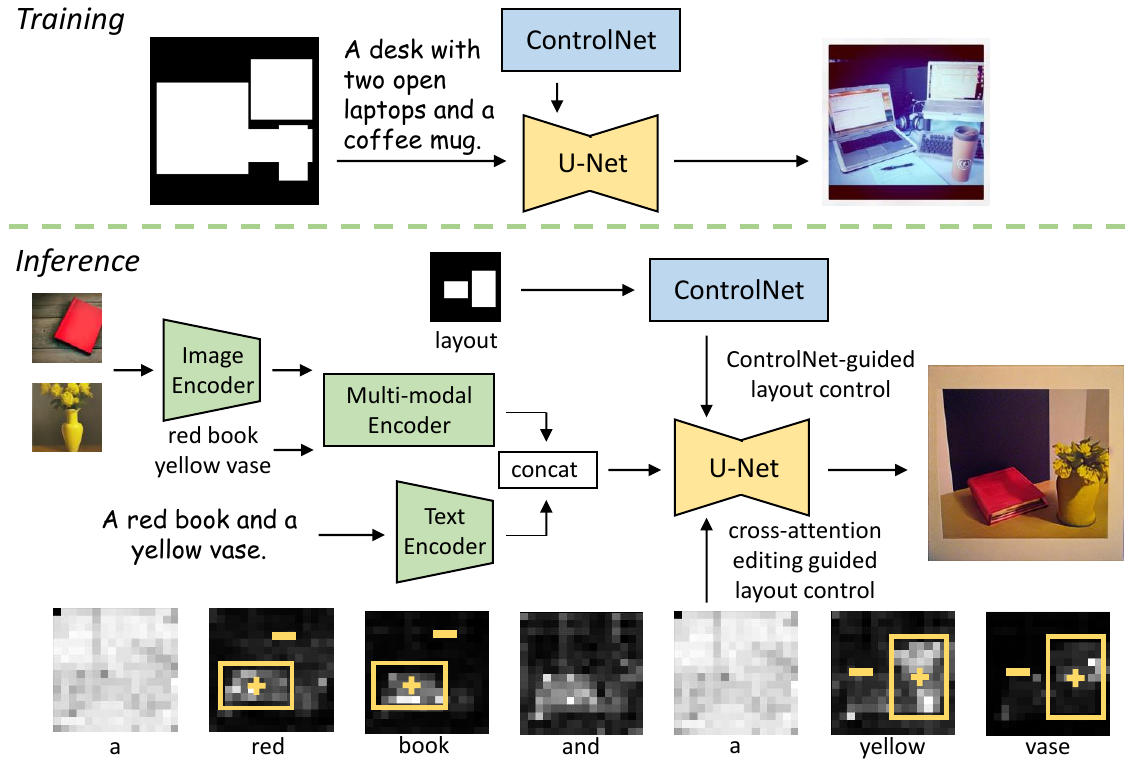}
   \caption{\textbf{Illustration of our tuning-free multi-concept customization tool.} ControlNet and cross-attention editing control the scene layout. The embeddings of multiple image concepts and the text prompt are concatenated together and forwarded into U-Net for image generation.}
   \label{fig:cus}
\end{figure}

In this section, we mainly introduce our tuning-free multi-concept customization tool. Its overview is illustrated in Fig.~\ref{fig:cus}. Training a tuning-free customization image generation model typically requires large-scale pre-training for subject representation learning. Currently, there are already tuning-free methods available that support single-concept customization. For our approach, we build upon the existing single-concept customization model, BLIP-Diffusion~\cite{li2023blip}, and extend its capability to accommodate multiple concepts, aided by the incorporation of the scene layout. Remarkably, we eliminate the need for large-scale upstream pre-training and directly construct a tuning-free multi-concept customization model to uphold the integrity of object attributes.

Specifically, for each concept, we extract its subject prompt embedding with the BLIP-2 encoders and the multi-modal encoder~\cite{li2023blipf}. To gain a comprehensive understanding of the precise object attributes, we harness the information contained within multiple images corresponding to a single concept. We collect all the embeddings derived from these images and compute their average, generating the definitive subject prompt embedding for subsequent use. They are concatenated with the text prompt embedding and forwarded into the U-Net of the single-concept customization model~\cite{li2023blip} for image generation.

However, directly aggregating embeddings from multiple concepts can easily lead to interference between different objects during image generation, resulting in the issue of concept confusion. To avoid this, we leverage the scene layout to regulate the positioning of each object, thereby mitigating the risk of their interference. We employ two levels of layout control - globally and locally. As is seen in the top part of Fig.~\ref{fig:cus}, we mask the background of the COCO dataset~\cite{lin2014microsoft}, and train a ControlNet~\cite{zhang2023adding} via a layout-to-image paradigm. The ControlNet is utilized to control the U-Net via residuals. It provides strong control at the global level, effectively distinguishing multiple objects, thus well avoiding their confusion. However, it can only control globally and cannot independently control the position of each individual object.

For local layout control of individual objects separately, we further propose to edit the cross-attention map according to the scene layout, motivated by the fact that the cross-attention map directly affects the spatial layout of the generated image~\cite{hertz2023prompttoprompt}. Specifically, we collect the cross-attention map of each object words and their attribute words. We add a positive constant number $\alpha^+$ to the regions corresponding to the presence of the object, while concurrently adding a negative constant number $\alpha^-$ to the rest regions. Compared to ControlNet, cross-attention editing realizes significantly weaker layout control but can independently govern the position of each object. Therefore, when synergistically integrated with ControlNet, it effectively controls the overall layout of the entire image. Ultimately, guided by the layout, different objects can be distinguished from each other, avoiding the confusion problem and achieving multi-concept customization.

\subsection{Layout-to-Image Generation}

To guarantee object relationships, we generate images directly from the scene layout. While our previously employed ControlNet and cross-attention editing approach can indeed tackle the layout-to-image problem, they are characterized by the imposition of too strong constraints upon the layout.  Once the scene layout deviates from the precise depiction of object relationships, it becomes challenging to guarantee the accurate representation of these relationships. Therefore, we utilize the strategy of latent updating by backwarding the box-constrained loss~\cite{xie2023boxdiff} for image generation from the layout. It provides a relatively loose control over the layout, thus allowing a flexible assurance of object relationships.

\subsection{Local Image Editing}

\begin{figure}
   \centering
\includegraphics[width=0.8\columnwidth]{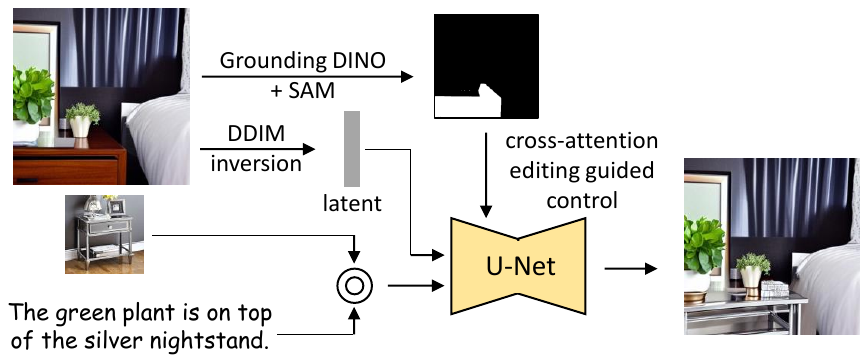}
   \caption{\textbf{Illustration of our local image editing tool.} The concept images pass into the customization network for embedding extraction, and the masked segmentation map guides the image generation process.}
   \label{fig:editing}
\end{figure}

To rectify objects with incorrect attributes, we introduce our local image editing tool, as illustrated in Fig.~\ref{fig:editing}. By querying our LLM agent for verification, we can identify which object attributes are erroneous and require modification. We utilize the combination of Grounding DINO~\cite{liu2023grounding} and SAM~\cite{kirillov2023segment} to segment the object out. The resulting segmented mask is utilized for cross-attention editing to provide position guidance for image editing. The image requiring editing is reconverted into the latent through DDIM inversion, serving as the initial latent for the subsequent image generation process. Images featuring objects characterized by correct attributes have already been generated earlier. These images, together with the text prompts, are processed in a manner analogous to the previous customization model, which serves as the conditional input for the U-Net. The process of image generation generally follows the previous multi-concept customization, with the image DDIM inversion as the initial latent. The segmentation masks are used as the guidance for cross-attention editing, while ControlNet is not employed. In this way, objects with incorrect attributes can be effectively substituted and rectified.

\section{Results}

We mainly conduct experiments on the recent T2I-CompBench benchmark~\cite{huang2023t2i}, which mainly divides compositional text prompts into six sub-categories. For quantitative comparison, we following its setting, and utilize the BLIP-VQA metric for attribute binding evaluation, the UniDet-based metric for spatial relationship evaluation, CLIPScore for non-spatial relationship, and the 3-in-1 metric for complex prompts. We also include qualitative results in both the main paper and supplementary materials.

\subsection{Quantitative Comparison}

\begin{table} 
\centering
\setlength{\abovecaptionskip}{0pt}
\setlength{\belowcaptionskip}{0pt}
\caption{\textbf{Quantitative Comparison on T2I-CompBench with existing text-to-image generation models and compositional methods}. Our method demonstrates superior compositional generation ability in both attribute binding, object relationships, and complex compositions. The best value is bolded, and the second-best value is underlined.} 
\label{tab:t2icompbench}
\resizebox{\columnwidth}{!}{ 
\begin{tabular}
{lcccccc}
\toprule
\multicolumn{1}{c}
{\multirow{2}{*}{\bf Model}} & \multicolumn{3}{c}{\bf Attribute Binding } & \multicolumn{2}{c}{\bf Object Relationship} & \multirow{2}{*}{\bf Complex$\uparrow$}
\\
\cmidrule(lr){2-4}\cmidrule(lr){5-6}

&
{\bf Color $\uparrow$ } &
{\bf Shape$\uparrow$} &
{\bf Texture$\uparrow$} &
{\bf Spatial$\uparrow$} &
{\bf Non-Spatial$\uparrow$} &
\\
\midrule
Stable v1.4  & 0.3765 & 0.3576 & 0.4156 & 0.1246 & 0.3079 & 0.3080  \\
Stable v2 & 0.5065 & 0.4221 & 0.4922 & 0.1342 & 0.3096 & 0.3386  \\
DALL-E 2 & 0.5750 & 0.5464 & 0.6374 & 0.1283 & 0.3043 & 0.3696  \\
SDXL & 0.6369 & 0.5408 & 0.5637 & 0.2032 & 0.3110 & 0.4091 \\
PixArt-$\alpha$ & 0.6886 & 0.5582 & 0.7044 & 0.2082 & 0.3179 & \underline{0.4117}  \\
DALL-E 3 & \underline{0.8110} & \underline{0.6750} & \textbf{0.8070} & - & - & - \\
\midrule
Composable Diffusion & 0.4063 & 0.3299 & 0.3645 & 0.0800 & 0.2980 & 0.2898  \\
Attn-Mask-Control & 0.4119 & 0.4649 & 0.4505 & 0.1249 &  0.3046 & 0.3779\\
StructureDiffusion & 0.4990 & 0.4218 & 0.4900 & 0.1386 & 0.3111 & 0.3355  \\
TokenCompose & 0.5055  & 0.4852  & 0.5881 & 0.1815  & 0.3173  & 0.2937 \\
Attn-Exct & 0.6400 & 0.4517 & 0.5963 & 0.1455 & 0.3109 & 0.3401  \\
GORS  & 0.6603 & 0.4785 & 0.6287 & 0.1815 & \underline{0.3193} & 0.3328  \\
ECLIPSE & 0.6119 & 0.5429  &0.6165  & 0.1903 & 0.3139 & - \\
LMD & 0.4838 & 0.5266 & 0.6215 &  \underline{0.4594} & 0.2735 & 0.3827\\
\midrule
\textbf{CompAgent (ours)} & \textbf{0.8488} &  \textbf{0.7233} & \underline{0.7916} & \textbf{0.4837} & \textbf{0.3212} & \textbf{0.4863} \\
\bottomrule
\end{tabular}
}
\vspace{-1em}
\end{table}

We list the quantitative metric results of our CompAgent in Tab.~\ref{tab:t2icompbench}. We compare with existing state-of-the-art text-to-image synthesis methods and models that are designed for complex text prompts. For text-to-image generation, we compare with the recent Stable Diffusion~\cite{rombach2022high} v1.4, v2 and XL~\cite{podell2023sdxl} model, DALL-E 2~\cite{ramesh2022hierarchical}, PixArt-$\alpha$~\cite{chen2023pixart} and DALL-E 3~\cite{betker2023improving}. For compositional text-to-image generation methods, we compare with Composable Diffusion~\cite{liu2022compositional}, StructureDiffusion~\cite{feng2023trainingfree}, Attn-Mask-Control~\cite{wang2023compositional}, GORS~\cite{huang2023t2i}. TokenCompose~\cite{wang2023tokencompose}, Attn-Exct~\cite{chefer2023attend}, ECLIPSE~\cite{patel2023eclipse} and LMD~\cite{lian2023llm} target at multiple objects within a sentence so we also compare with them. 

For attribute binding, our method achieves significantly higher BLIP-VQA metric compared to previous methods, 16.02\%, 16.51\% and 8.72\% higher for the color, shape and texture attributes compared with PixArt-$\alpha$ respectively. Compared to DALL-E 3, the current state-of-the-art method for controllability in text-to-image generation, our approach performs on par with it in terms of the performance. For the color and shape attributes, our CompAgent is even 3.78\% and 4.83\% higher than it, based on the Stable Diffusion model. This demonstrates the capability of CompAgent to accurately generate object types as well as their attributes. Compared to single-step generation, the superiority of such the divide-and-conquer multi-step generation manner can thus be observed in attribute binding. For object relationships, CompAgent excels in both spatial and non-spatial relationships. In contrast, previous methods either lack such ability or are limited to handling only a single type of relationship. With the ability for both attribute binding and object relationship, our CompAgent can well address complex text prompts: we achieve the 48.63\% 3-in-1 metric, which surpasses previous methods by more than 7\%. The quantitative results well demonstrate that our method effectively addresses the challenges associated with compositional text-to-image generation.

\subsection{Ablation Study}

\begin{table} 
\centering
\setlength{\abovecaptionskip}{0pt}
\setlength{\belowcaptionskip}{0pt}
\caption{\textbf{Ablation study on T2I-CompBench about LLM agent planning and verification}. "Customization" denotes our tuning-free multi-concept customization tool, and "layout-to-image" denotes our layout-to-image generation tool. } 
\label{tab:planning}
\resizebox{0.94\columnwidth}{!}{ 
\begin{tabular}
{lcccc}
\toprule
\multicolumn{1}{c}
{\bf Model} & 
{\bf Color $\uparrow$ } &
{\bf Shape$\uparrow$} &
{\bf Texture$\uparrow$} &
{\bf Complex $\uparrow$} \\
\midrule
customization &  0.8160 & 0.6839 & 0.7692 &  0.4184\\
layout-to-image & 0.6626 & 0.5505 & 0.6707 &  0.4387\\
+ planning & 0.8458 & 0.7211 & 0.7916 & 0.4642 \\
+ verification & \textbf{0.8488} (\textbf{{\color{gre} +3.28\%}})&  \textbf{0.7233} (\textbf{{\color{gre}+3.94\%}}) & \textbf{0.7916} (\textbf{{\color{gre}+2.24\%}}) & \textbf{0.4863} (\textbf{{\color{gre}+4.76\%}}) \\
\bottomrule
\end{tabular}
}
\vspace{-1em}
\end{table}

\begin{figure*}
\centering
\includegraphics[width=0.99\textwidth]{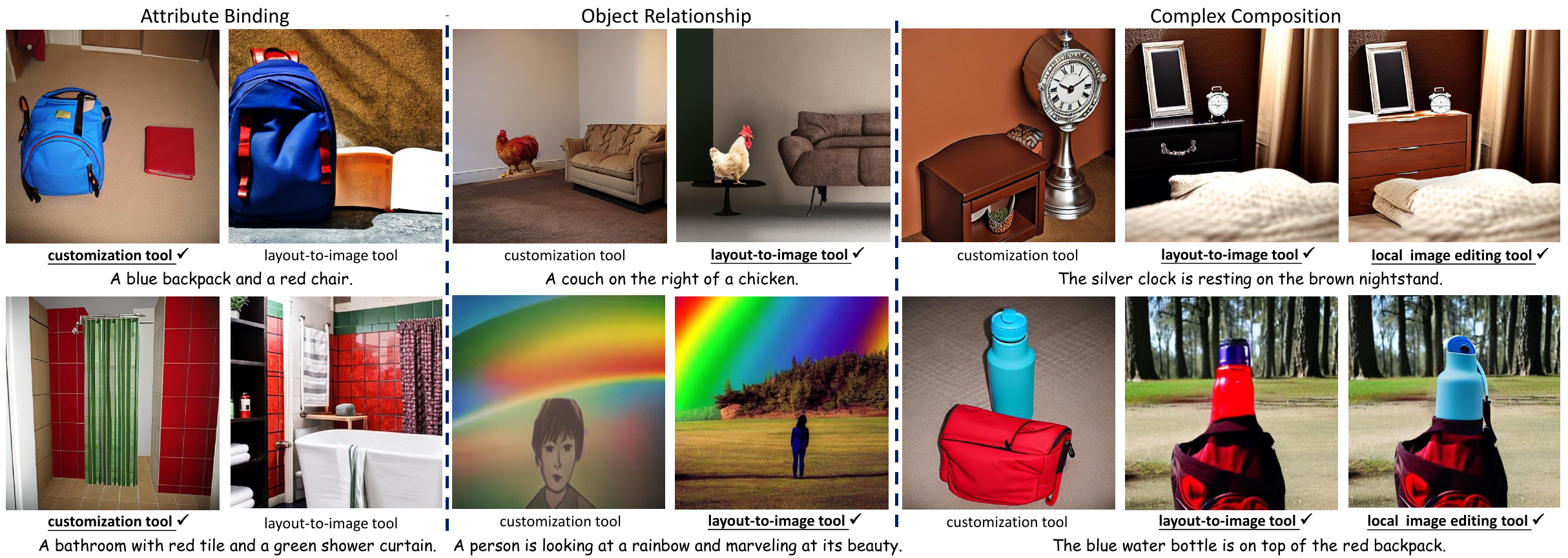}
\setlength{\abovecaptionskip}{0pt}
\setlength{\belowcaptionskip}{0pt}
  \caption{\textbf{Visualized examples of CompAgent to show the generated images from different tools and how LLM agent plans to use the tools.} The LLM agent analyzes the text prompt. It employs the customization tool to address attribute binding, and the layout-to-image tool to address object relationships. For complex composition, the layout-to-image tool is utilized for object relationship, then the local image editing tool is used for attribute correction. }
  \label{fig:illuex}  
\end{figure*}

\begin{figure*}
\centering
\includegraphics[width=0.96\textwidth]{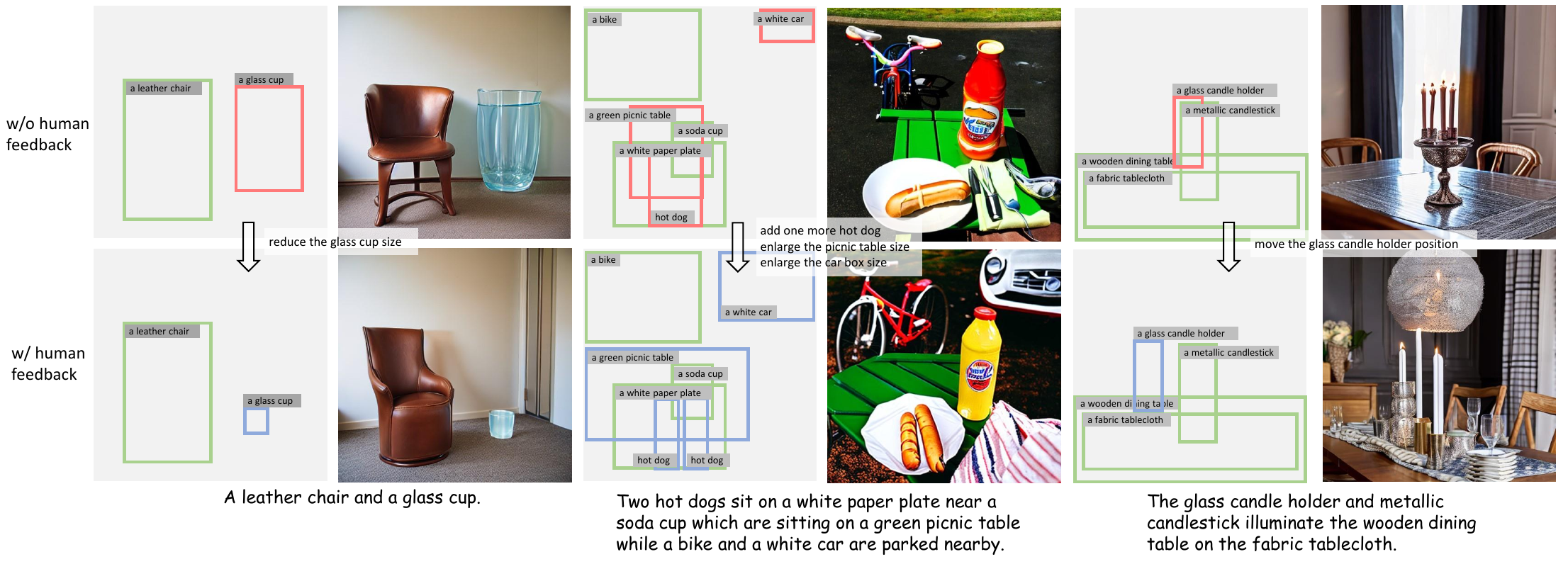}
\setlength{\abovecaptionskip}{0pt}
\setlength{\belowcaptionskip}{0pt}
  \caption{\textbf{Visualized examples to show the effect of human feedback.} The green boxes are well-generated by the LLM agent. The red boxes are the ones generated by the LLM agent with some issues, which are modified through human feedback into the blue boxes in the second row. }
  \label{fig:humanfeed}  
\end{figure*}

We further conduct ablation study in this section to analyze the effect of the LLM agent under our framework.

\textit{LLM agent planning and verification.} We first analyze the effect of the planning and verification mechanism of our LLM agent in Tab.~\ref{tab:planning}. It can be observed that by leveraging the individual object images, the multi-concept customization tool performs well for attribute binding. However, because of the utilized ControlNet, the customization model can be inflexible for expressing the object relationships, which leads to its limited metric scores in complex compositions. In comparison, the layout-to-image generation tool can well generate images with object relationships, but cannot guarantee the accuracy of object attributes. Our LLM agent can analyze the complex text prompts, and plan the most suitable tool to use. As a result, LLM agent planning well helps address most situations for compositional text-to-image generation. The verification mechanism of our LLM agent further helps correct some attribute errors, especially in complex compositions where layout-to-image generation cannot handle the object attributes, which thus contributes to the 2.21\% improvement. As we can see, by planning and verification, our LLM agent well utilizes the tools for compositional text-to-image generation.

We provide some visualized examples in Fig.~\ref{fig:illuex}. As we can see, the LLM agent well plans and employs the most suitable tool. The customization tool is utilized to strictly constrain object attributes, and the layout-to-image tool can make appropriate adjustments to the layout for object relationships. Besides, the local image editing tool can assist in rectifying objects with incorrect attributes.

\textit{Human feedback.} We further provide some visualized examples in Fig.~\ref{fig:humanfeed} to show the effect of human feedback. For the first example, although object types and attributes are correct, the size of the glass cup is too large. By involving human feedback to modify the scene layout, such a problem can be addressed. For the second example, the texts are quite complex and there are some mistakes in the scene layout - one less hot dog, small table and too small car. Humans can inspect and correct them, then CompAgent can generate accurate images. This also applies to the third example. Our CompAgent can incorporate human feedback, enabling it to generate more realistic images and handle more complex text prompts.

\subsection{Qualitative Comparison}

\begin{figure*}
\centering
\includegraphics[width=0.9\textwidth]{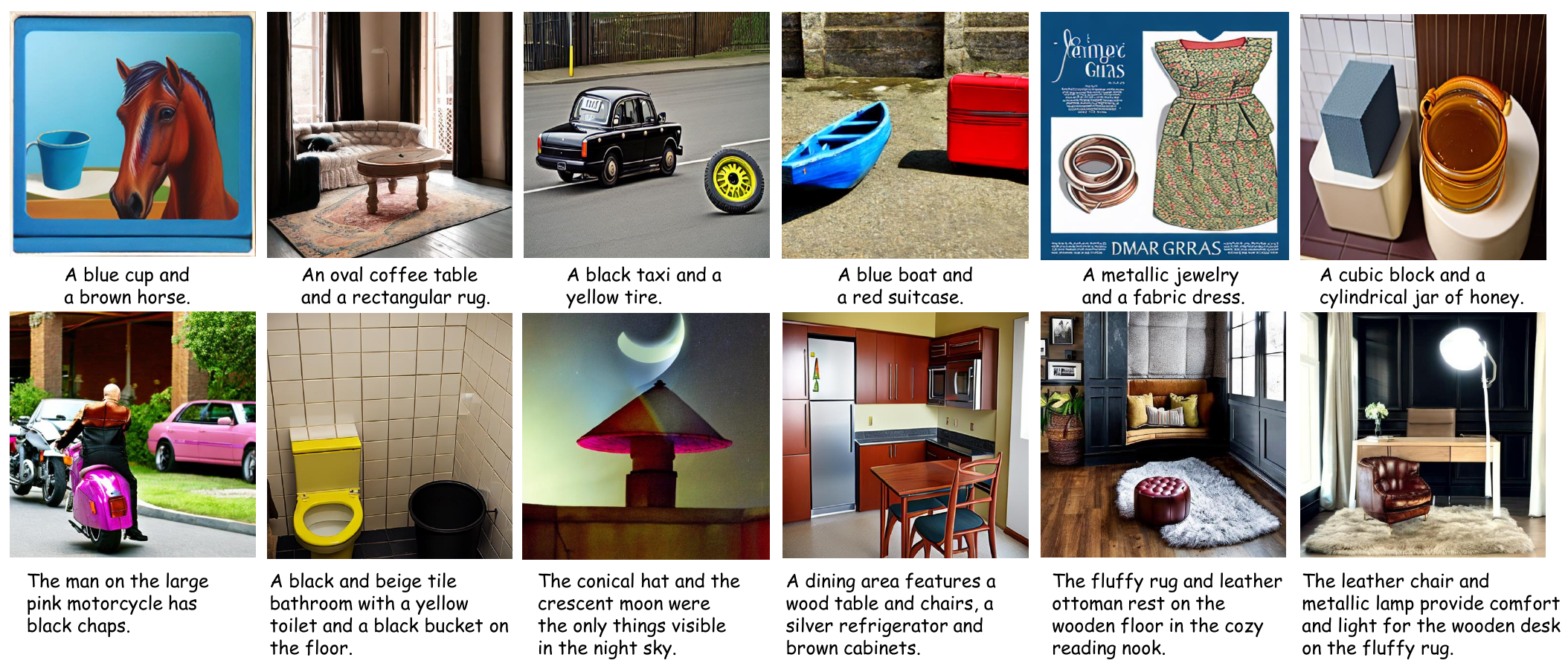}
\setlength{\abovecaptionskip}{0pt}
\setlength{\belowcaptionskip}{0pt}
  \caption{\textbf{Visualized results of our method for attribute binding.} }
  \label{fig:rcolor}  
\end{figure*}

\begin{figure*}
\centering
\includegraphics[width=0.9\textwidth]{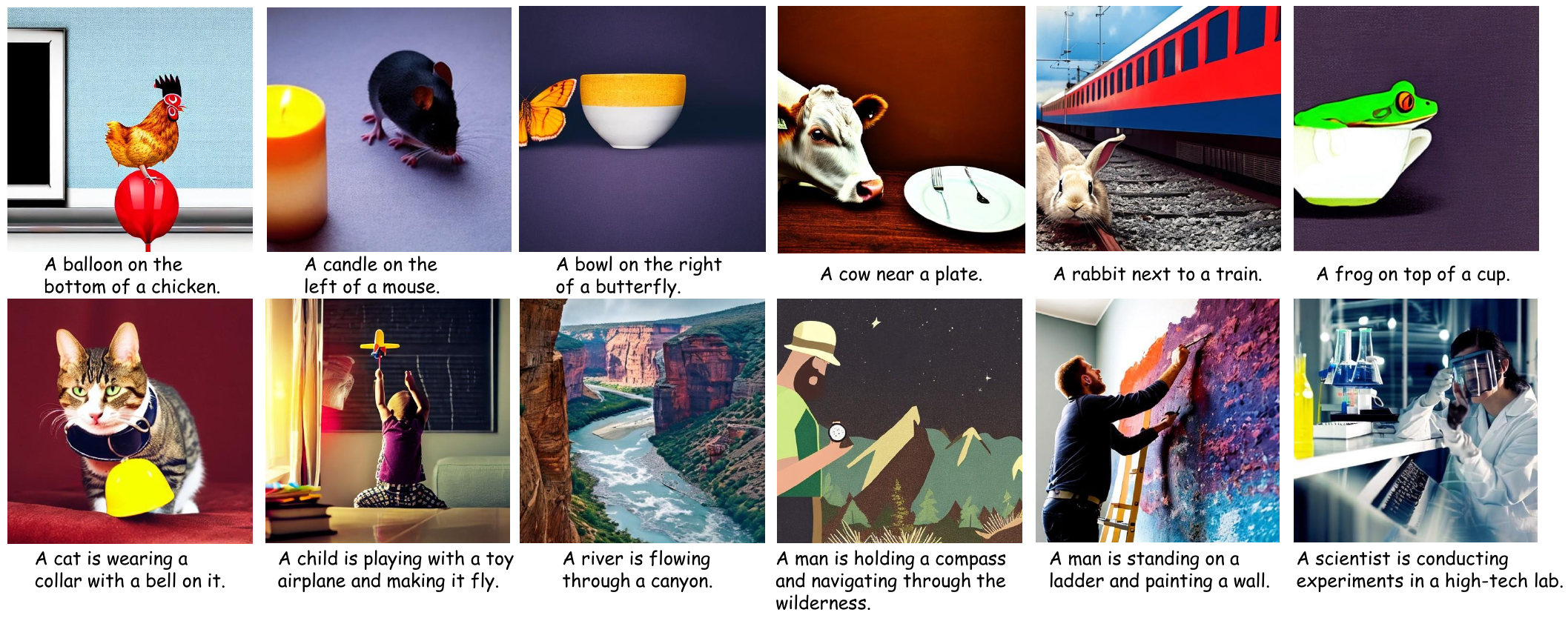}
\setlength{\abovecaptionskip}{0pt}
\setlength{\belowcaptionskip}{0pt}
  \caption{\textbf{Visualized results of our method for object relationship.} }
  \label{fig:rrelation}  
\end{figure*}

\begin{figure*}
\centering
\includegraphics[width=0.9\textwidth]{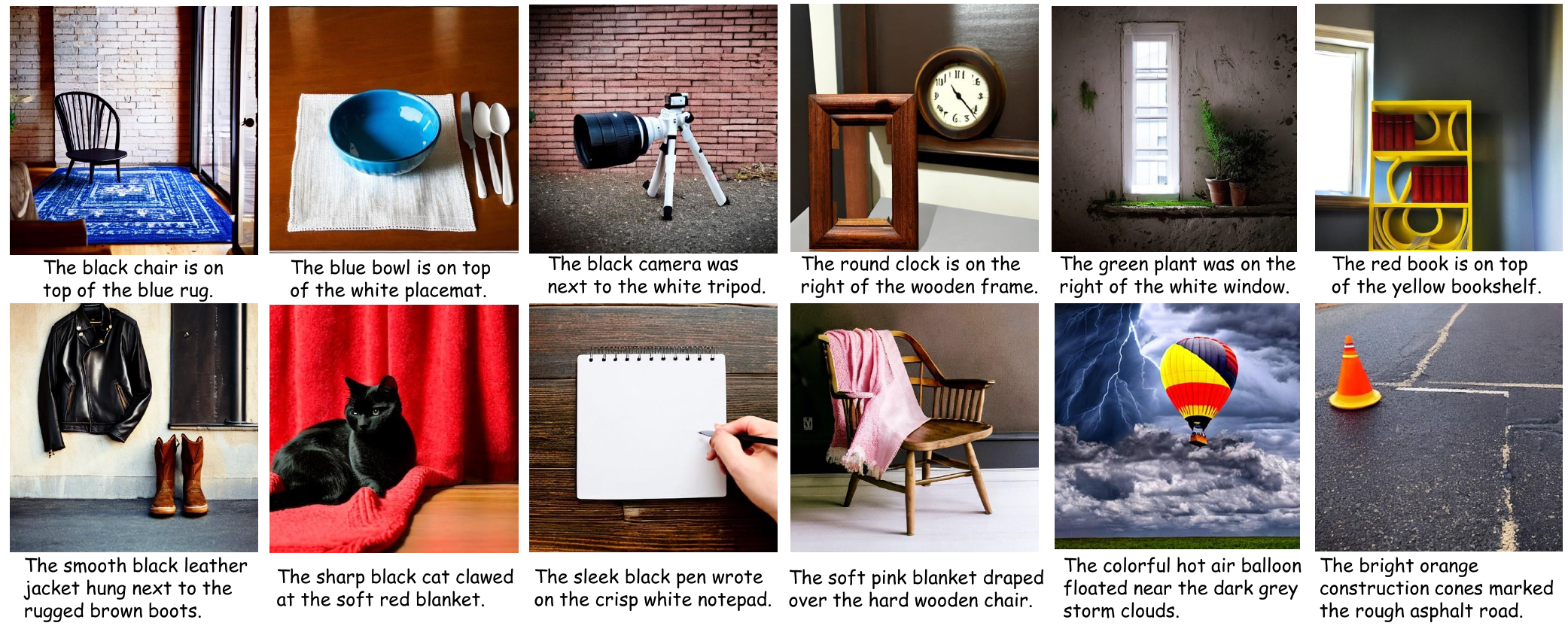}
\setlength{\abovecaptionskip}{0pt}
\setlength{\belowcaptionskip}{0pt}
  \caption{\textbf{Visualized results of our method for complex composition.} }
  \label{fig:rcomplex}  
\end{figure*}

We provide more visualized results in Fig.~\ref{fig:rcolor}, Fig.~\ref{fig:rrelation}, Fig.~\ref{fig:rcomplex}. CompAgent generates correctly for object types, attributes and relationships. The excellent ability of our method for compositional text-to-image generation can thus be further demonstrated.

\begin{figure*}
\includegraphics[width=0.99\textwidth]{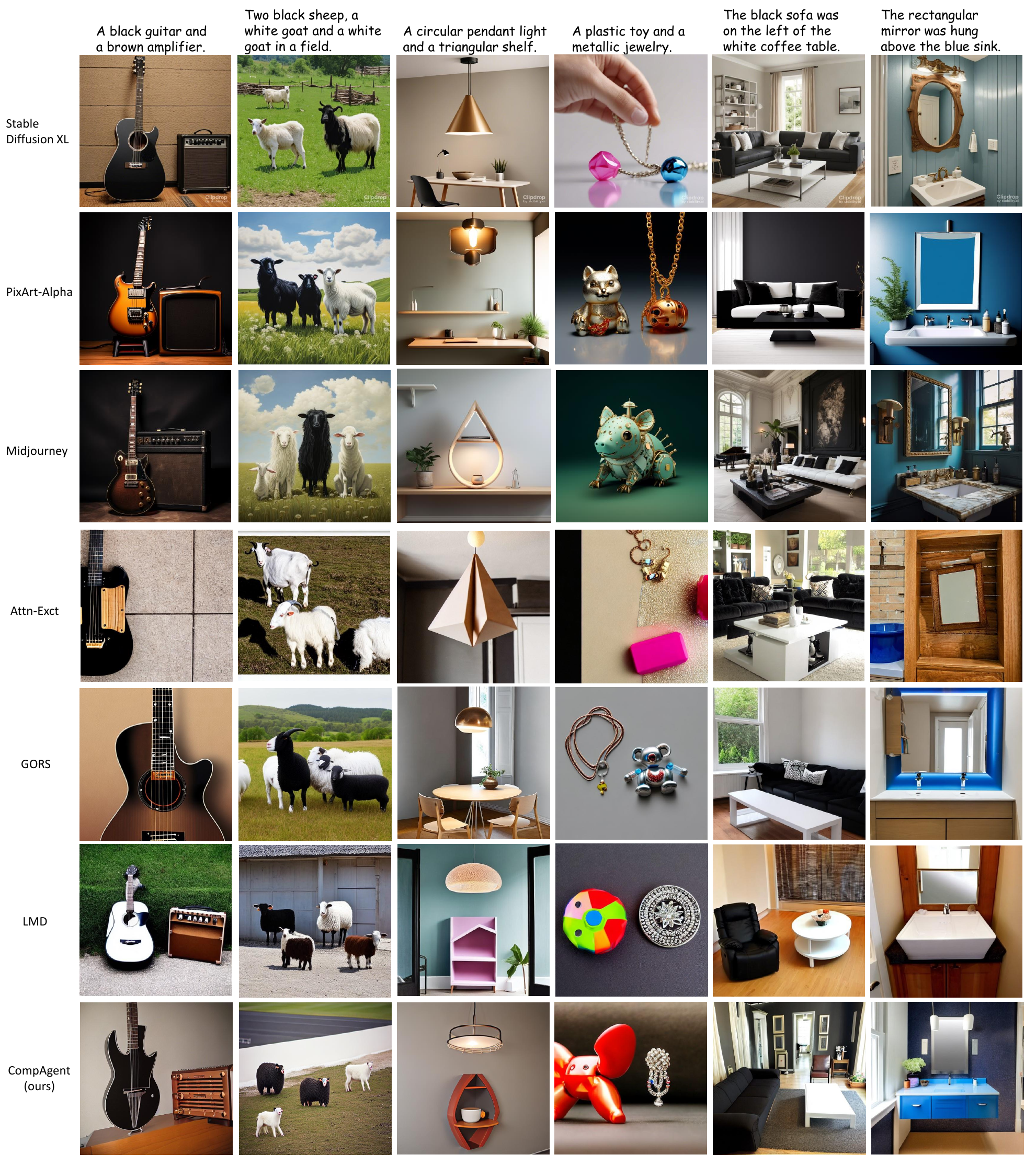}
  \caption{\textbf{Qualitative comparison between our approach and previous methods.} The first three lines are existing text-to-image generation models, and lines four to six are existing compositional text-to-image generation methods. Our approach performs significantly better in both attribute binding and object relationship aspects compared to previous methods.}
  \label{fig:comparison2}  
\end{figure*}

We then provide the qualitative comparison with existing text-to-image generation methods and compositional text-to-image generation methods in Fig.~\ref{fig:comparison2}. For the text "a black guitar and a brown amplifier", existing methods are easy to confuse the color of the guitar and the amplifier. In the second example, where four objects exist in the text, the correct object number also cannot be guaranteed for existing methods. For some uncommon attributes, like the triangular shelf in the third example and the blue sink in the last example, existing models are also easy to make mistakes about the attributes. Besides, in the fifth example, most of existing methods cannot express the "left" relationship accurately. Our CompAgent generates accurately for all these text prompts. This further demonstrates the superiority of our method over existing models when it comes to compositional text-to-image generation.

\section{Conclusion}

In this paper, we propose a training-free approach, CompAgent, for compositional text-to-image generation. By decomposing, planning, verifying, and involving human feedback, the LLM agent coordinates the whole system and employs external tools to generate high-fidelity and accurate images according to the given complex text prompts. Extensive results demonstrate that CompAgent well addresses the object type and quantity, attribute binding, and object relationship problems in compositional text-to-image generation. We consider CompAgent as an important step towards the future of autonomous agents empowered by language models and the controllability in text-to-image generation.

%
%
%
%



\newpage
\appendix

\section{Implementation Details}

The main part of our experiments applies GPT-4 \cite{gpt42023} as our LLM agent. For our toolkits, the multi-modal models utilized for attribute verification are implemented by GPT-4V. The image generation tools are implemented based on Stable Diffusion \cite{rombach2022high}. The text-to-image generation tool and the layout-to-image generation tool are based on Stable Diffusion v2, and the tuning-free multi-concept customization tool and the local image editing tool are based on Stable Diffusion v1-4. For training ControlNet \cite{zhang2023adding} in the layout-to-image manner, we utilize the COCO dataset \cite{lin2014microsoft} and train for 10 epochs. $\alpha^+$ is set to 2.5 and $\alpha^-$ is set to -10,000 for cross attention editing.

\section{Agent Prompt}

Our prompt input for the LLM agent mainly contains the following parts:

\noindent 1. Task specification:

\begin{tcolorbox}[colback={colorcommentbg}, colframe={colorcommentframe}, left=3pt, right=3pt, top=3pt, bottom=3pt, boxrule=1pt]
You are an intelligent research assistant. You have access to the following tools to address the compositional text-to-image generation problem:
\end{tcolorbox}

\noindent 2. Tool library introduction:

\begin{tcolorbox}[colback={colorcommentbg}, colframe={colorcommentframe}, left=3pt, right=3pt, top=3pt, bottom=3pt, boxrule=1pt]

1) \textit{The multi-concept customization tool}. It is good at texts where objects are coupled complex attributes, like color, shape, texture. It can also handle certain spatial relationships, like “on the left of, on the right of” and some simple and straightforward object relationships, like the mirror hanging above, the snow covering.

2) \textit{The layout-to-image generation tool}. It is good at texts where objects are interacted with complicated relationships or actions, like playing, holding. It can also process attributes that are easy or straightforward, like a red apple, the white snow.

3) \textit{The text-to-image generation tool}. It can handle those simple texts without complex attributes and relationships.

\end{tcolorbox}

\noindent 3. Decomposition details:

\begin{tcolorbox}[colback={colorcommentbg}, colframe={colorcommentframe}, left=3pt, right=3pt, top=3pt, bottom=3pt, boxrule=1pt]
I will provide you with a caption for an image. Your should also extract individual objects and generate the bounding boxes for the objects mentioned in the caption. The images are of size 512x512. The top-left corner has coordinate [0, 0]. The bottom-right corner has coordinate [512, 512]. The bounding boxes should not go beyond the boundaries and it is better to avoid overlapping bounding boxes. Each bounding box should be in the format of (object name, [top-left x coordinate, top-left y coordinate, box width, box height]) and include exactly one object (i.e., start the object name with a or an if possible). If needed, you can make reasonable guesses. Please refer to the example below for the desired format.
\end{tcolorbox}

\noindent 4. In-context examples:

\begin{tcolorbox}[colback={colorcommentbg}, colframe={colorcommentframe}, left=3pt, right=3pt, top=3pt, bottom=3pt, boxrule=1pt]
$\mathcal{Q}_1$:

Caption: a blue horse and a brown vase.

$\mathcal{A}_1$:

Analysis: attribute-only.

Objects: [('a blue horse', [50, 70, 220, 300]), ('a brown vase', [300, 113, 150, 250])]

$\mathcal{Q}_2$:

Caption: a fabric rug and a leather belt.

$\mathcal{A}_2$:

Analysis: attribute-only.

Objects: [('a fabric rug', [20, 200, 470, 150]), ('a leather belt', [100, 250, 300, 20])]

$\mathcal{Q}_3$:

Caption: A cat is wearing a collar with a bell on it.

$\mathcal{A}_3$:

Analysis: relationship-only.

Objects: [('a cat', [120, 150, 300, 300]), ('a collar', [120, 300, 300, 50]), ('a bell, [250, 320, 110, 100])]

$\mathcal{Q}_4$:

Caption: The blue bowl was on top of the white placemat.

$\mathcal{A}_4$:

Analysis: both.

Objects: [('a rectangular mirror', [170, 80, 172, 100]), ('a white sink', [150, 200, 212, 150])]

$\mathcal{Q}_5$:

Caption: The red apple was on top of the plate.

$\mathcal{A}_5$:

Analysis: relationship-only.

Objects: [('a red apple', [235, 230, 60, 60]), ('a plate', [175, 210, 180, 180])]
\end{tcolorbox}

After providing the above prompt to the agent, we then prompt the LLM agent for completion:

\begin{tcolorbox}[colback={colorcommentbg}, colframe={colorcommentframe}, left=3pt, right=3pt, top=3pt, bottom=3pt, boxrule=1pt]
Caption: [input prompt from the user]
\end{tcolorbox}

The resulting text analysis and decomposition results from the LLM agent is then parsed and used for the subsequent image generation process.

\section{FURTHER ABLATIONS AND EXPERIMENTS}

\subsection{Agent Type}

\begin{table}[h]
\centering
\setlength{\abovecaptionskip}{0pt}
\setlength{\belowcaptionskip}{0pt}
\caption{\textbf{The impact of LLM agent type on T2I-CompBench for compositional text-to-image generation}. } 
\label{tab:type}
\resizebox{\columnwidth}{!}{ 
\begin{tabular}
{lcccccc}
\toprule
\multicolumn{1}{c}
{\multirow{2}{*}{\bf LLM Agent}} & \multicolumn{3}{c}{\bf Attribute Binding } & \multicolumn{2}{c}{\bf Object Relationship} & \multirow{2}{*}{\bf Complex$\uparrow$}
\\
\cmidrule(lr){2-4}\cmidrule(lr){5-6}

&
{\bf Color $\uparrow$ } &
{\bf Shape$\uparrow$} &
{\bf Texture$\uparrow$} &
{\bf Spatial$\uparrow$} &
{\bf Non-Spatial$\uparrow$} &
\\
\midrule
Llama-7B  & 0.6994 & 0.5740 & 0.6927  &  0.3138 & 0.3102 & 0.3515\\
Llama-70B &  0.7400 & 0.6305 & 0.7102 & 0.3698 & 0.3104 & 0.4475 \\
GPT-3.5 & 0.7925 &  0.6647 & 0.7743 & 0.4439 &  0.3195 & 0.4752\\
GPT-4 & \textbf{0.8488} &  \textbf{0.7233} & \textbf{0.7916} & \textbf{0.4837} & \textbf{0.3212} & \textbf{0.4863} \\
\bottomrule
\end{tabular}
}
\vspace{-1em}
\end{table}

We first conduct compositional text-to-image generation experiments on the T2I-CompBench \cite{huang2023t2i} guided by the agent implemented by various LLMs. The experimental results are listed in Tab. \ref{tab:type}. We adopt GPT-3.5, GPT-4 and the open-sourced Llama 2 \cite{touvron2023llama2}.  

It can be seen that as the language understanding ability of large language models improves, the performance of our CompAgent for compositional text-to-image generation also enhances. When adopting GPT-4 as the agent, the metric reaches the highest. This is because GPT-4 can not only decompose individual objects successfully, but also generate the scene layout where the shapes of bounding boxes are more suitable for the object types. Meanwhile, even utilizing the Llama-7B model as our agent, the compositional text-to-image generation metric is still higher than most of existing methods. This demonstrates the flexibility of our CompAgent across different large language models.

\subsection{Image Generation with Layout Guidance}

\begin{table}[t]
\centering
\setlength{\abovecaptionskip}{0pt}
\setlength{\belowcaptionskip}{0pt}
\caption{\textbf{Evaluation for layout-to-image generation}. We adopt the YOLO score to evaluate the correspondence of images and their layouts. } 
\label{tab:layout}
\resizebox{0.85\columnwidth}{!}{ 
\begin{tabular}
{lccc}
\toprule
Methods & AP $\uparrow$ & AP$_{50}$ $\uparrow$ & AP$_{75}$ $\uparrow$ \\
\midrule
LostGAN  & 0.053 & 0.089 & 0.056\\
LAMA  & 0.102 & 0.153 & 0.117\\
TwFA  & 0.106 & 0.147 & 0.126\\
Stable Diffusion  & 0.028  & 0.092  & 0.011 \\
GLIGEN  &  0.297  & 0.458  & 0.339\\
GLIGEN + BoxDiff & 0.402  & 0.620  & 0.462 \\
\midrule
latent updating  & 0.224 & 0.468 & 0.178\\
cross-attention editing & 0.060 & 0.190 & 0.021\\
ControlNet & 0.338 & 0.521 & 0.339\\
ControlNet + cross-attention editing & \textbf{0.508} & \textbf{0.778} & \textbf{0.534} \\
\bottomrule
\end{tabular}
}
\vspace{-1em}
\end{table}

For composing multiple individual objects into the image, the generated scene layouts guide the generation process under our CompAgent framework. We compare the performance of different layout-to-image generation methods and list the results in Tab. \ref{tab:layout}. We compare with previous methods, including LostGAN \cite{sun2021learning}, LAMA \cite{li2021image}, TwFA \cite{yang2022modeling}, Stable Diffusion \cite{rombach2022high}, GLIGEN \cite{li2023gligen} and GLIGEN + BoxDiff \cite{xie2023boxdiff}. Under our CompAgent framework, we mainly utilize latent updating, cross-attention editing and ControlNet three strategies. We conduct experiments on the benchmark proposed in \cite{xie2023boxdiff}. We apply the YOLOv4 \cite{bochkovskiy2020yolov4} to detect objects and obtain the YOLO score, including AP, AP$_{50}$ and AP$_{75}$ to evaluate the precision of the layout-to-image performance.

It can be seen that by utilizing cross-attention editing only to control image generation through bounding boxes, the obtained YOLO score is quite low, only 0.06, slightly higher than Stable Diffusion. This demonstrates that although editing cross-attention maps can provide layout guidance, it is still insufficient to control image generation accurately. Utilizing our trained ControlNet is much more effective, achieving the 0.338 AP, 4.1\% higher than GLIGEN. However, ControlNet can only provide global control, not object-level. After combining ControlNet and cross-attention editing, the YOLO score reaches to 0.508, 10.6\% higher than GLIGEN + BoxDiff. This demonstrates that our design can control the positions of objects more accurately. This well avoids the confusion of multiple objects, thus is suitable for the attribute binding problem. However, the reliance on the scene layout is too strong in this way, making it inappropriate for generating correct object relationships. Instead, latent updating can well follow the scene layout and can also be flexible meanwhile. Therefore, we utilize the latent updating strategy for the object relationship issue.

\subsection{Qualitative Comparison}

\begin{figure*}
\centering
\includegraphics[width=0.99\textwidth]{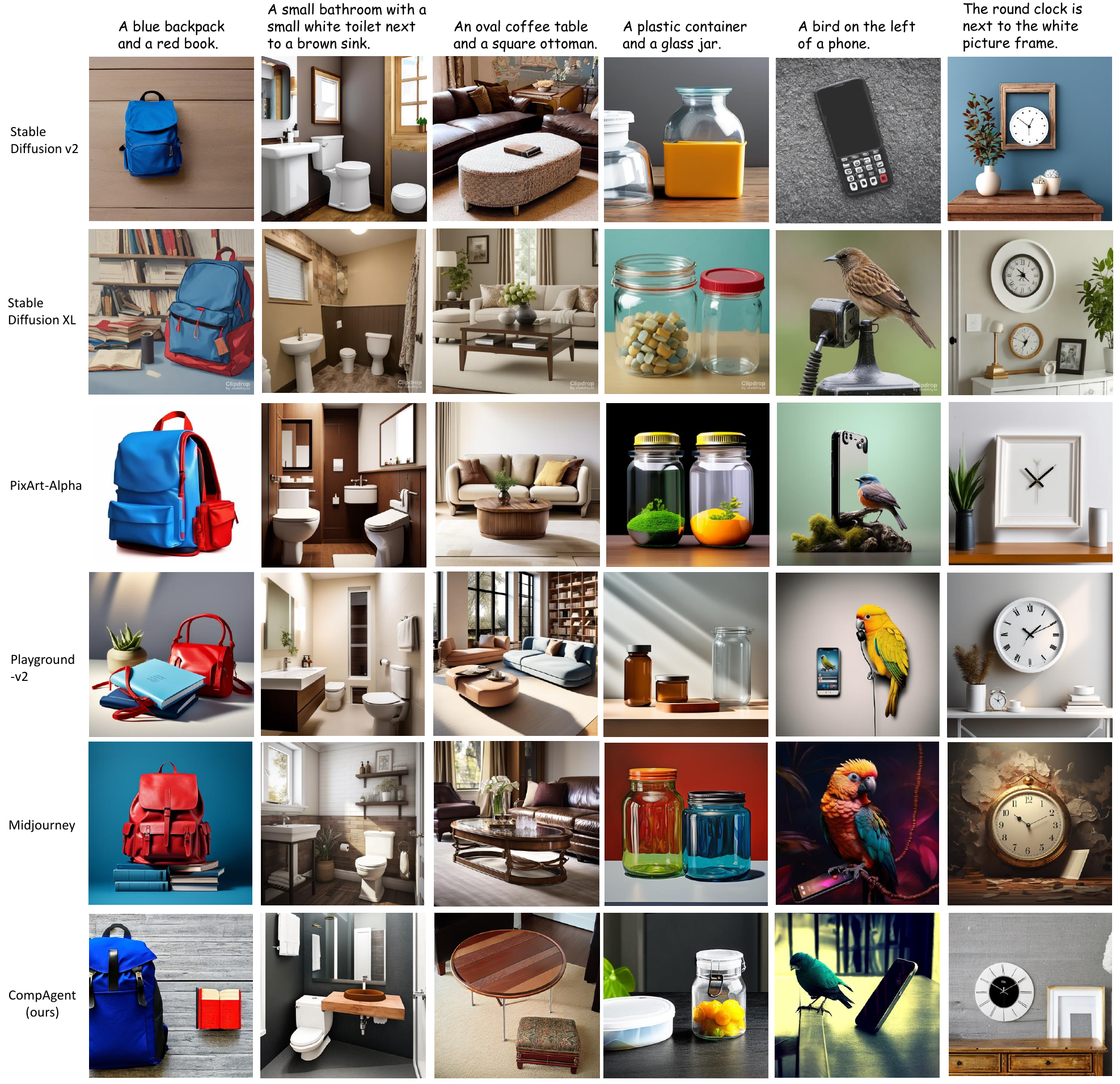}
\setlength{\abovecaptionskip}{0pt}
\setlength{\belowcaptionskip}{0pt}
  \caption{\textbf{Qualitative comparison between our approach and existing state-of-the-art text-to-image generation methods.} }
  \label{fig:supp_comp1}  
\end{figure*}

\begin{figure*}
\centering
\includegraphics[width=0.99\textwidth]{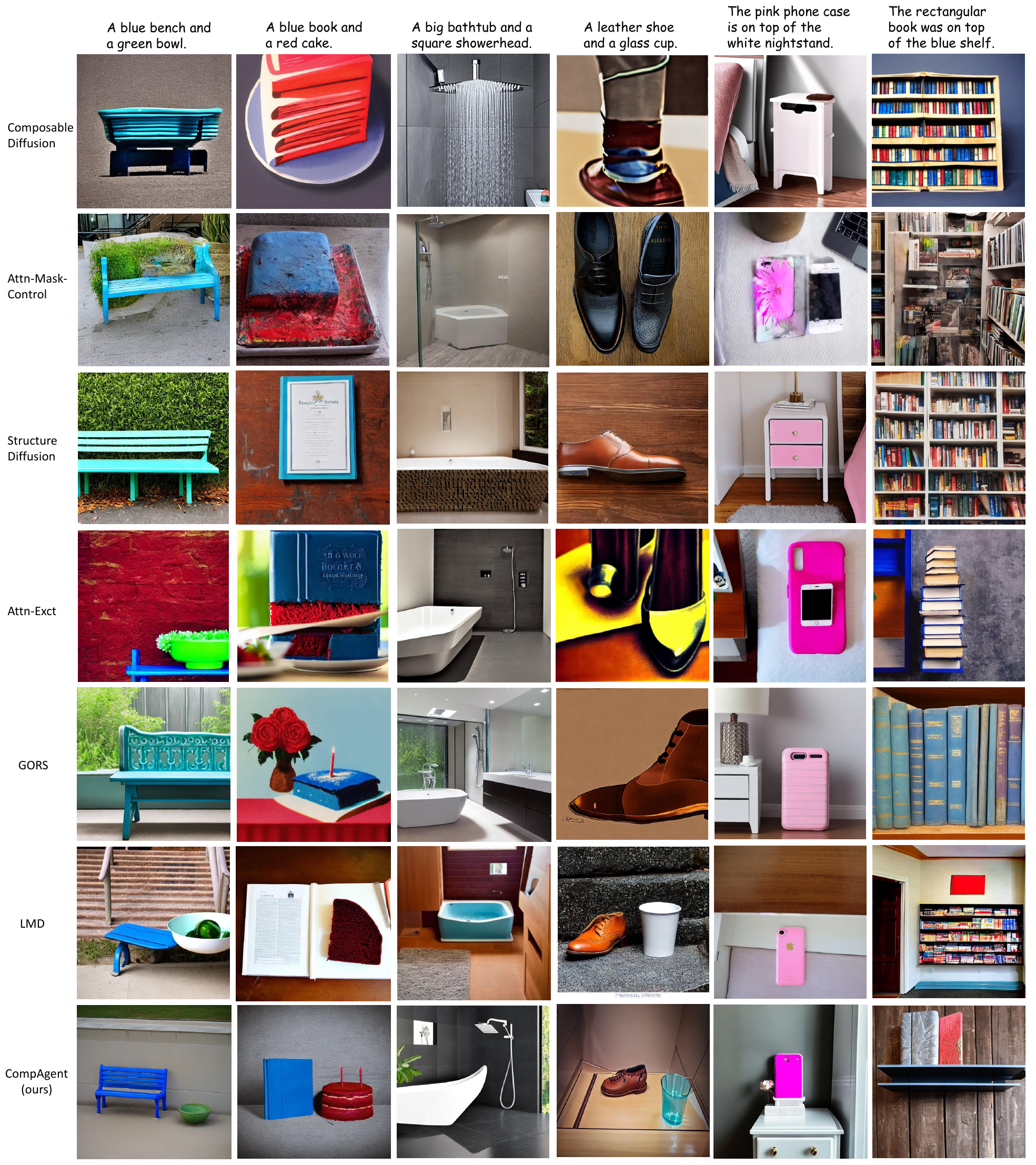}
\setlength{\abovecaptionskip}{0pt}
\setlength{\belowcaptionskip}{0pt}
  \caption{\textbf{Qualitative comparison between our approach and existing compositional text-to-image generation methods.} }
  \label{fig:supp_comp2}  
\end{figure*}

We then provide more visualized comparisons with existing state-of-the-art text-to-image generation methods in Fig. \ref{fig:supp_comp1}. Existing methods are highly prone to the following errors. 1) Attribute confusion. For example, for the text "a blue backpack and a red book", existing models are easy to confuse the color of the backpack and the red book, or generate a backpack with red parts. 2) Constrained by common attributes or scenarios. For example, for the second example, existing models tend to generate a white sink since it is more common in reality, rather than the brown one. For the third example, existing methods also tend to generate a common living room scenario, while ignoring the required objects and attributes. 3) Incorrect relationship. For example, the "left" relationship cannot be correctly expressed in the fifth example. In comparison, our CompAgent well avoids these problems, thus generates images that more accurately align with the description of input texts.

We also provide visualized comparison with existing compositional text-to-image generation methods in Fig. \ref{fig:supp_comp2}. We can observe that existing compositional text-to-image methods also cannot address the above mentioned problems. As a result, the correct object types and quantities, object attributes and relationships cannot be guaranteed. CompAgent behaves equally well for these examples. 

\begin{figure*}
\centering
\includegraphics[width=0.9\textwidth]{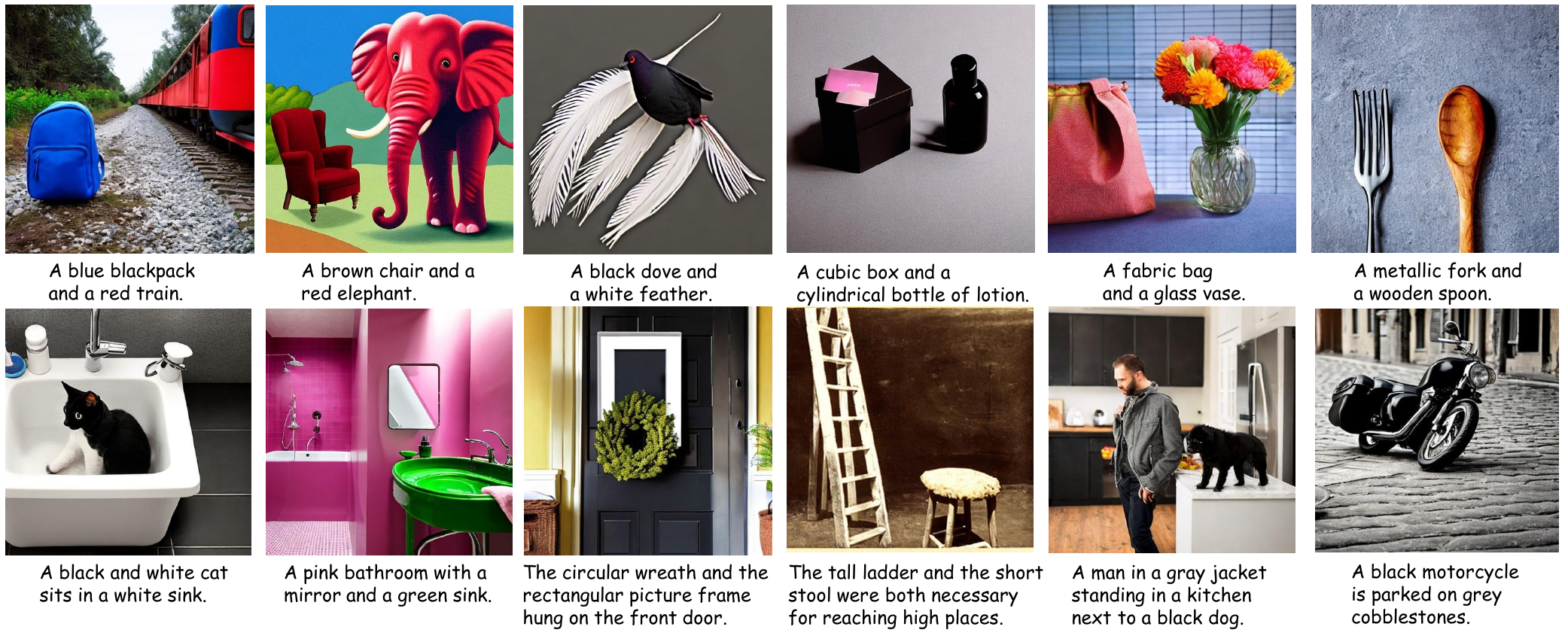}
\setlength{\abovecaptionskip}{0pt}
\setlength{\belowcaptionskip}{0pt}
  \caption{\textbf{Visualized results of our method for attribute binding.} }
  \label{fig:supp_rcolor}  
\end{figure*}

\begin{figure*}
\centering
\includegraphics[width=0.9\textwidth]{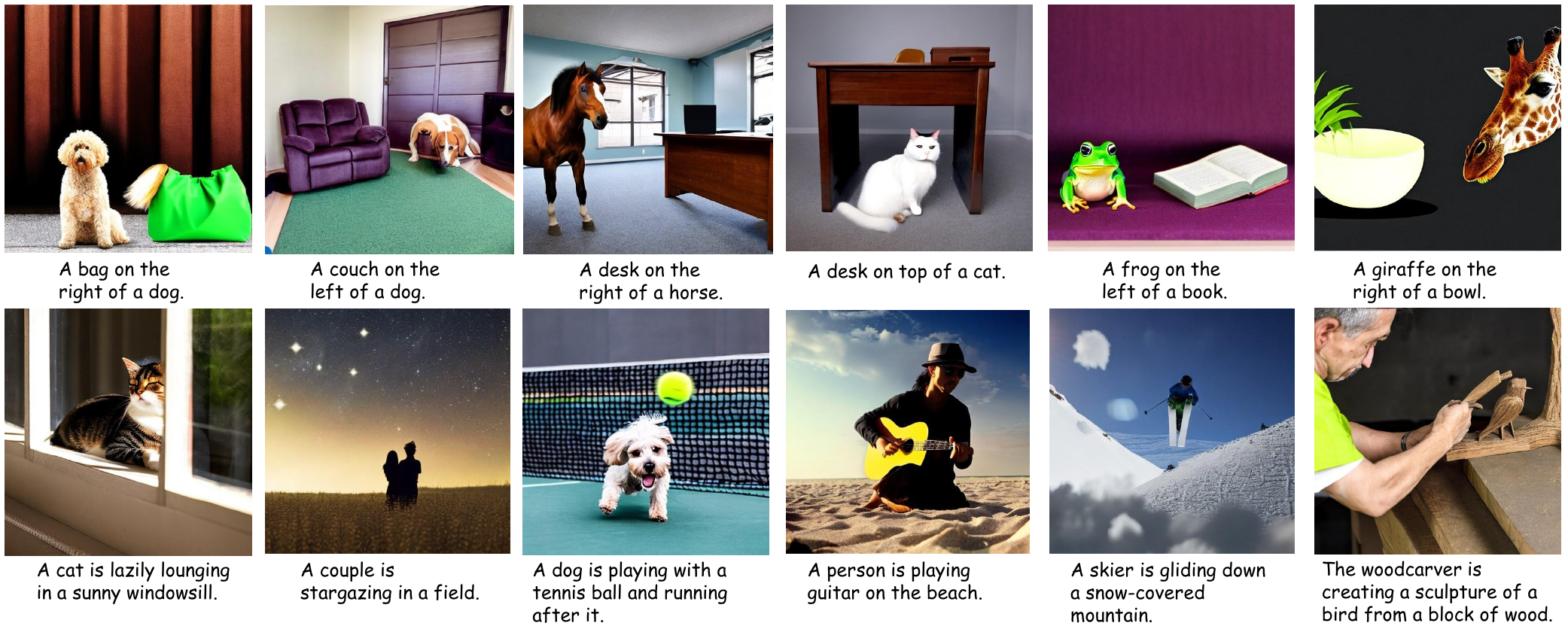}
\setlength{\abovecaptionskip}{0pt}
\setlength{\belowcaptionskip}{0pt}
  \caption{\textbf{Visualized results of our method for object relationship.} }
  \label{fig:supp_rrelation}  
\end{figure*}

\begin{figure*}
\centering
\includegraphics[width=0.9\textwidth]{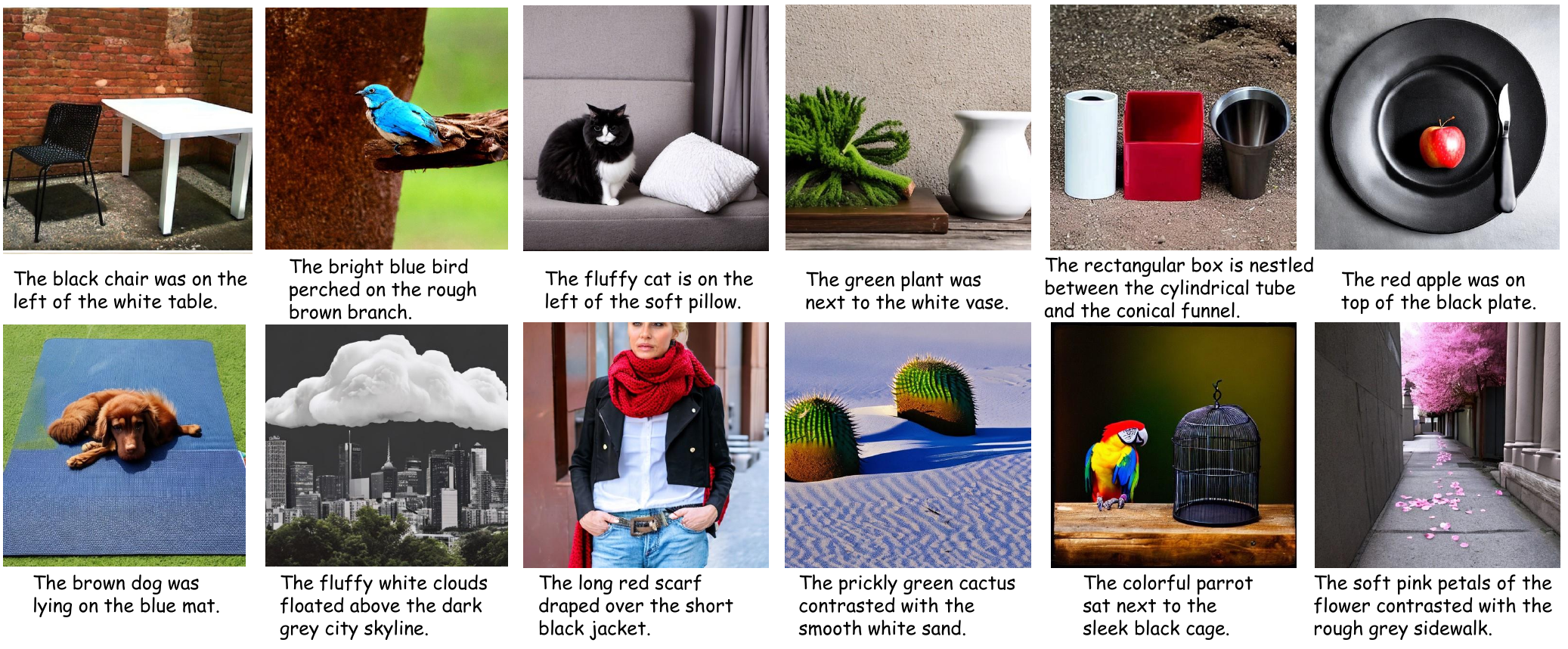}
\setlength{\abovecaptionskip}{0pt}
\setlength{\belowcaptionskip}{0pt}
  \caption{\textbf{Visualized results of our method for complex composition.} }
  \label{fig:supp_rcomplex}  
\end{figure*}

More visualized examples are provided in Fig. \ref{fig:supp_rcolor}, Fig. \ref{fig:supp_rrelation} and Fig. \ref{fig:supp_rcomplex}. These examples further demonstrate the excellent ability of our CompAgent for addressing the compositional text-to-image generation problem.

\section{Extension to other tasks}

Besides the compositional text-to-image generation task, our CompAgent can also be flexibly extended to other related image generation tasks with the help of the LLM agent and our toolkits. We mainly introduce about the multi-concept customization, the image editing and the object placement tasks.

\begin{figure*}
\centering
\includegraphics[width=0.9\textwidth]{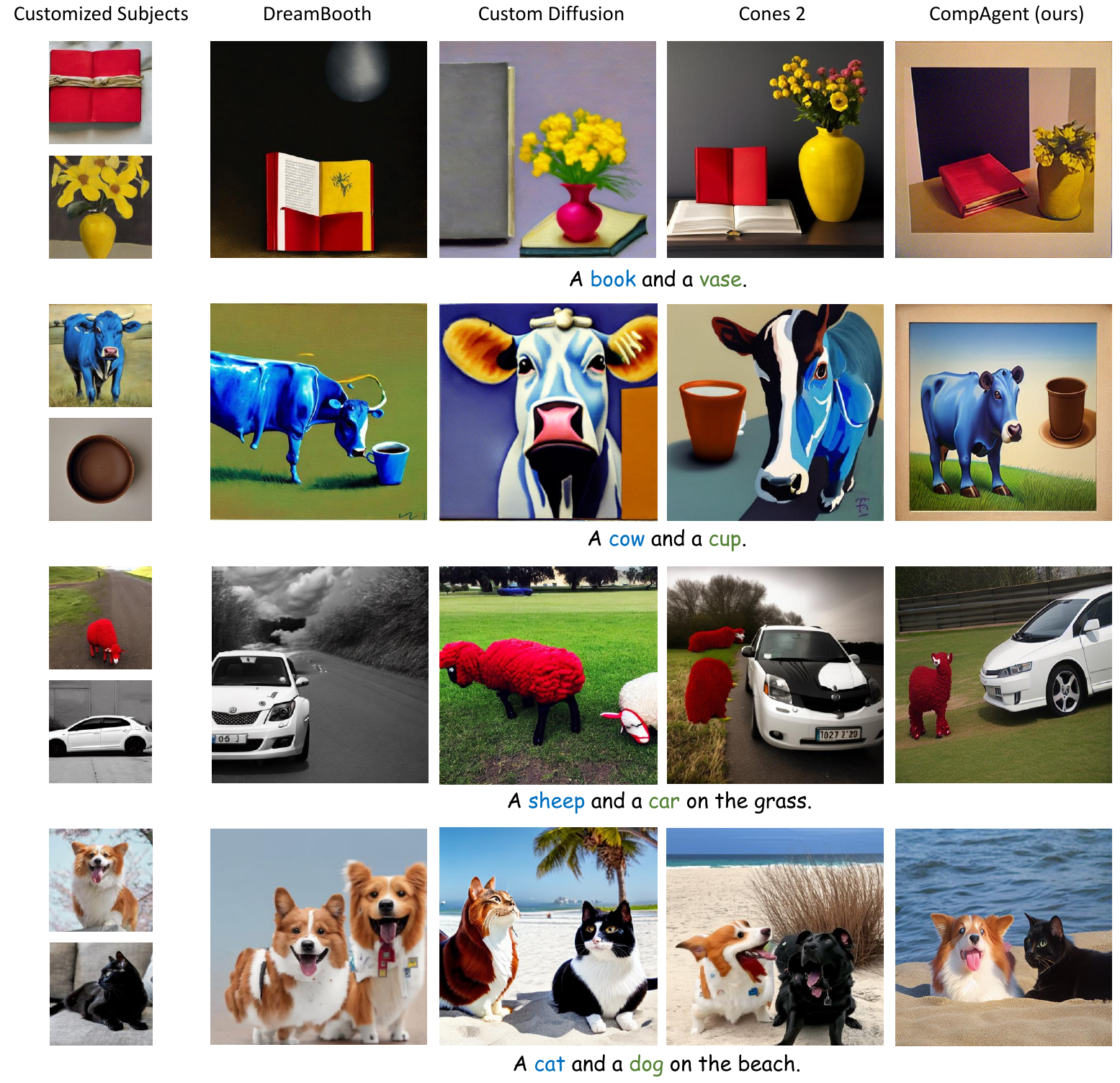}
\setlength{\abovecaptionskip}{0pt}
\setlength{\belowcaptionskip}{0pt}
  \caption{\textbf{Visualized results of our method for multi-concept customization.} Note that DreamBooth, Custom Diffusion and Cones 2 are all tuning-based methods, while our CompAgent is tuning-free.}
  \label{fig:rcus}  
\end{figure*}

\subsection{Multi-Concept Customization}

We first conduct the multi-concept customization task to generate images according to the input text prompts containing the given subjects. We compare with existing state-of-the-art customization methods, including DreamBooth \cite{ruiz2023dreambooth}, Custom Diffusion \cite{kumari2023multi} and Cones 2 \cite{liu2023cones}. The comparison results are provided in Fig. \ref{fig:rcus}. It can be seen that DreamBooth and Custom Diffusion cannot generate the corresponding objects or their correct attributes. For example, for the first example, DreamBooth does not generate the vase in the image, and the object attributes from the Custom Diffusion image are incorrect. For the second example, Custom Diffusion does not generate the cup, while DreamBooth generates the cup with the incorrect color. Cones 2 performs better than them, generating accurate images with a red book and a yellow vase. However, it is also limited by common attributes. For example, it does not generate the correct color for the cow in the second example, which also applies to the car in the third example. Besides, in the fourth example, Cones 2 confuses the features of the cat and the dog, thus generates the image with two dogs. In comparison, our CompAgent accurately captures the subject features and avoids the object confusion problem, thus handles the multi-concept customization task better. Note that these previous methods are all tuning-based, while our CompAgent is tuning-free. Therefore, CompAgent can accurately address the customization task more efficiently.

\begin{figure*}
\centering
\includegraphics[width=0.9\textwidth]{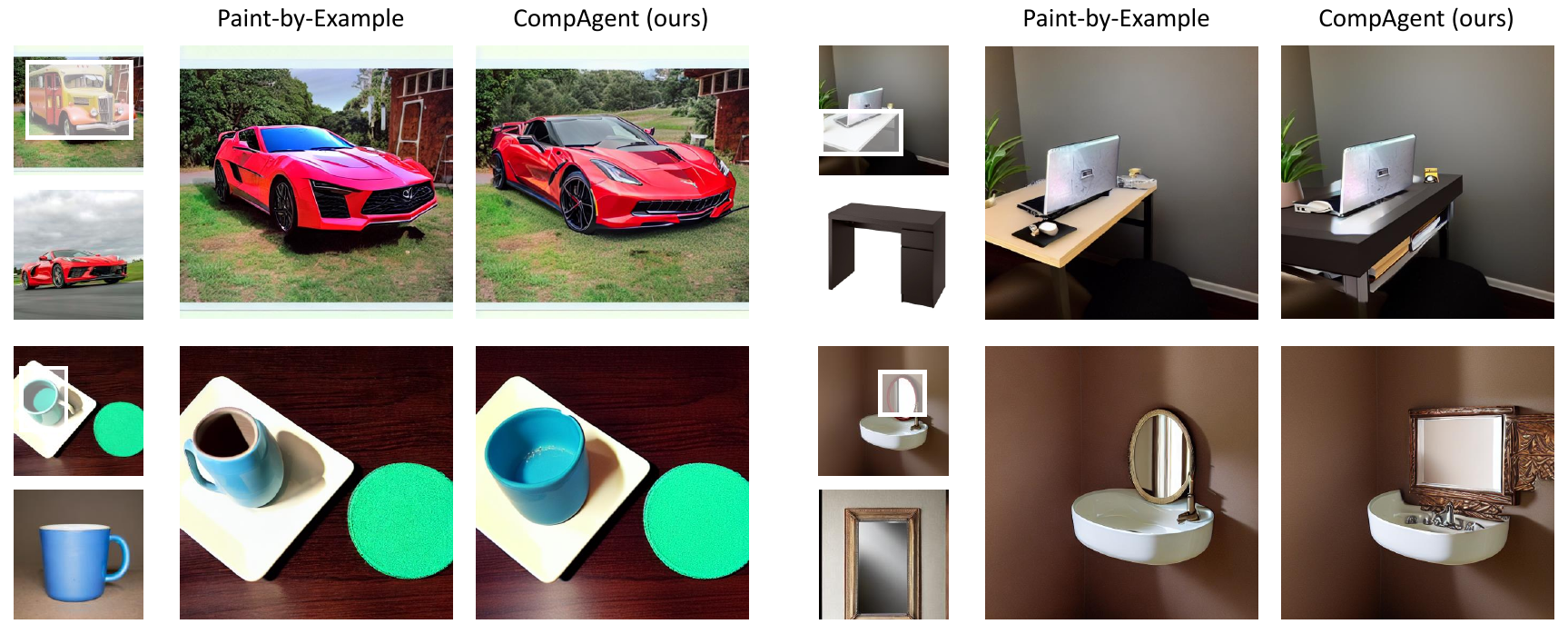}
\setlength{\abovecaptionskip}{0pt}
\setlength{\belowcaptionskip}{0pt}
  \caption{\textbf{Visualized results of our method for local image editing.} }
  \label{fig:redit}  
\end{figure*}

\begin{figure*}
\centering
\includegraphics[width=0.9\textwidth]{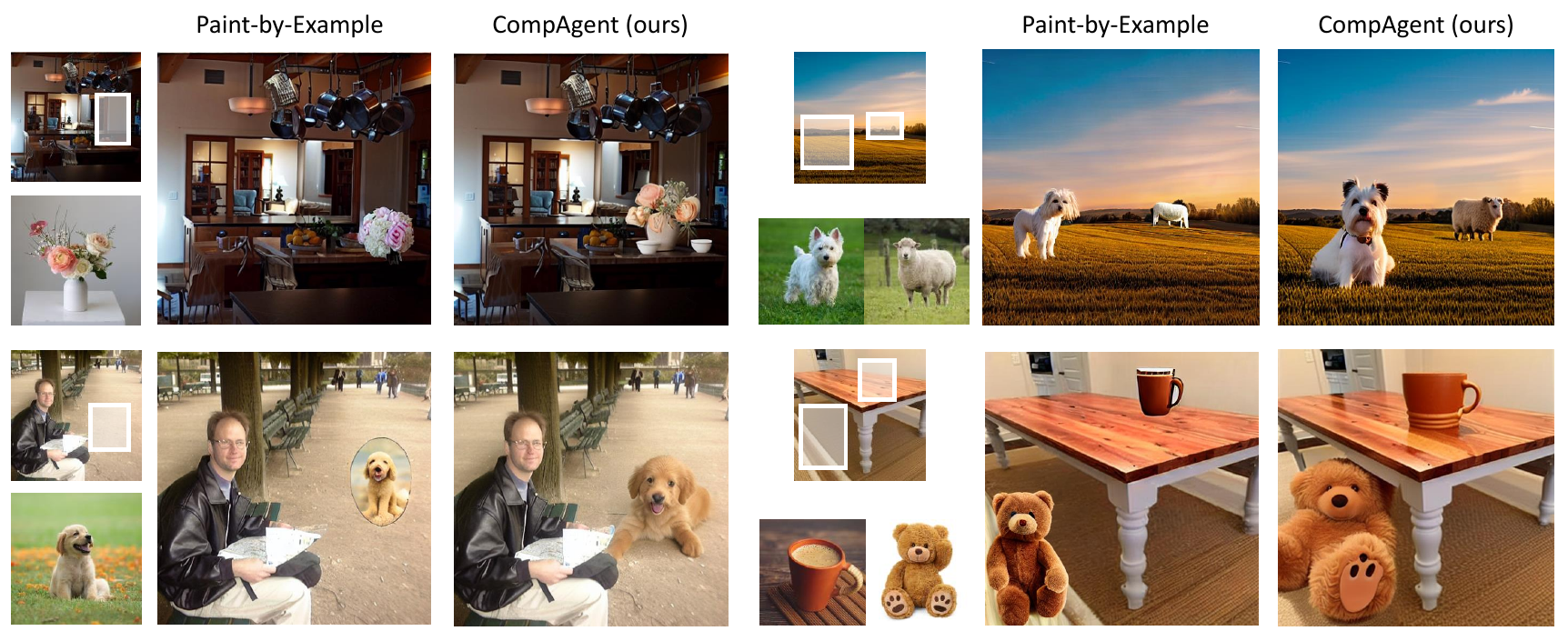}
\setlength{\abovecaptionskip}{0pt}
\setlength{\belowcaptionskip}{0pt}
  \caption{\textbf{Visualized results of our method for object placement.} }
  \label{fig:rplace}  
\end{figure*}

\subsection{Local Image Editing}

We then conduct the local image editing experiments and compare with the Paint-by-Example method \cite{yang2023paint}. It can be seen that although Paint-by-Example can perform the local image editing task, it cannot precisely catch the object attributes. For example, for the car example, the front window color from the Paint-by-Example generated image turns to blue. For the computer-desk example, Paint-by-Example does not edit the color of the table, and for the mirror-sink example, Paint-by-Example also does not modify the mirror shape from oval to rectangle. In comparison, our CompAgent can well guarantee the object attributes, thus better addressing the local image editing task. Besides, in the cup-coaster example, the shape of the cup from Paint-by-Example looks weird. In comparison, the generated image from our CompAgent looks more realistic.

\subsection{Object Placement}

Finally we conduct the object placement task: to put a given object into the specified position in an image. The comparison results with Paint-by-Example are illustrated in Fig. \ref{fig:rplace}. Our method performs equally better. For example, for the vase example, Paint-by-Example only generates the flowers and does not understand the object type. For the bear-cup example, Paint-by-Example does not capture the cup color. Besides, Paint-by-Example also does not generate natural images for the dog example. Our CompAgent performs better for all these examples, which further demonstrates its ability.


\bibliographystyle{ACM-Reference-Format}
\bibliography{sample-bibliography}

\end{document}